\documentclass[12pt]{article}
\usepackage{amsmath}
\usepackage{times}
\usepackage{graphicx}
\usepackage{color}
\usepackage{multirow}
\usepackage[authoryear]{natbib}
\usepackage{rotating}
\usepackage{bbm}
\usepackage{latexsym}

\usepackage{tikz}
\usepackage{float}
\usepackage{subcaption}
\usepackage{pgfplots}
\pgfplotsset{
	compat=newest,
}

\usepgfplotslibrary{external} 
\usetikzlibrary{patterns}
\usetikzlibrary{decorations.markings}
\makeatletter
\tikzset{
	nomorepostactions/.code={\let\tikz@postactions=\pgfutil@empty},
	mymark/.style 2 args={decoration={markings,
			mark= between positions 0 and 1 step (1/10)*\pgfdecoratedpathlength with{%
				\tikzset{#2,every mark}\tikz@options
				\pgfuseplotmark{#1}%
			},
		},
		postaction={decorate},
		/pgfplots/legend image post style={
			mark=#1,#2,every path/.append style={nomorepostactions}
		},
	},
}
\makeatother

\definecolor{brinkpink}{rgb}{0.98, 0.38, 0.5}
\definecolor{darkblue}{rgb}{0.0, 0.0, 0.55}
\definecolor{bluemat}{rgb}{0.26, 0.33, 0.55}
\definecolor{redmat}{rgb}{0.6, 0.25, 0.21}
\definecolor{greenmat}{rgb}{0.28, 0.63, 0.3}

\renewcommand{\thesubsubsection}{\arabic{section}.\arabic{subsubsection}}

\newcommand{\captionfonts}{\normalsize}

\makeatletter  
\long\def\@makecaption#1#2{%
	\vskip\abovecaptionskip
	\sbox\@tempboxa{{\captionfonts #1: #2}}%
	\ifdim \wd\@tempboxa >\hsize
	{\captionfonts #1: #2\par}
	\else
	\hbox to\hsize{\hfil\box\@tempboxa\hfil}%
	\fi
	\vskip\belowcaptionskip}
\makeatother   
\begin{document}
	\hspace{13.9cm}1
	
	\ \vspace{20mm}\\
	
	{\LARGE Exploring Tradeoffs in Spiking Neural Networks}
	
	\ \\
	{\bf \large Florian Bacho$^{\displaystyle 1}$ and Dominique Chu$^{\displaystyle 1}$}\\
	{$^{\displaystyle 1}$CEMS, School of Computing, University of Kent, Canterbury CT2 7NF, UK}\\
	%
	
	%\ \\[-2mm]
	
	\thispagestyle{empty}
	\markboth{}{NC instructions}
	\ \vspace{-0mm}\\
	%
	%Abstract
	\begin{center} {\bf Abstract} \end{center}
	Spiking Neural Networks (SNNs) have emerged as a promising alternative to traditional Deep Neural Networks for low-power computing. However, the effectiveness of SNNs is not solely determined by their performance but also by their energy consumption, prediction speed, and robustness to noise. The recent method Fast \& Deep, along with others, achieves fast and energy-efficient computation by constraining neurons to fire at most once. Known as Time-To-First-Spike (TTFS), this constraint however restricts the capabilities of SNNs in many aspects. In this work, we explore the relationships between performance, energy consumption, speed and stability when using this constraint. More precisely, we highlight the existence of tradeoffs where performance and robustness are gained at the cost of sparsity and prediction latency. To improve these tradeoffs, we propose a relaxed version of Fast \& Deep that allows for multiple spikes per neuron. Our experiments show that relaxing the spike constraint provides higher performance while also benefiting from faster convergence, similar sparsity, comparable prediction latency, and better robustness to noise compared to TTFS SNNs. By highlighting the limitations of TTFS and demonstrating the advantages of unconstrained SNNs we provide valuable insight for the development of effective learning strategies for neuromorphic computing.
	%%%%%%%%%%%

\section{Introduction}

Over the last decade, Deep Neural Networks (DNNs) have become indispensable tools in statistical machine learning, achieving state-of-the-art performance in various applications, including Computer Vision \citep{imagnet_dnn, dnn_for_object_detection}, Natural Language Processing \citep{attention_is_all_you_need, bert_dnn, gpt3}, and Reinforcement Learning \citep{atari_dqn, async_actor_critic}. However, their impressive performance often comes at a significant hardware and energy cost. For example, Natural Language Processing models can consist of billions of parameters and require energy-intensive GPU clusters to train efficiently \citep{gpt3}. These hardware and energy requirements pose a significant challenge in terms of sustainability and restrict the practical applicability of DNNs in resource-limited environments such as low-powered edge devices. Therefore, exploring more energy-efficient alternatives to DNNs is crucial not only to address the environmental cost of machine learning but also to provide practical and sustainable solutions in edge computating.
\par
One possible alternative to DNNs is Spiking Neural Networks (SNNs). Spiking neurons process information through discrete spatio-temporal events known as \textit{spikes} rather than continuous real-number values \citep{snn_third_generation, spiking_neuron_models_single_neuron_populations_plasticity}. Spikes enable efficient implementations of neural networks on non-von Neumann neuromorphic hardware such as Intel Loihi, IBM TrueNorth, Brainscale2, and SpiNNaker \citep{spinnaker, truenorth, training_in_the_loop_brainscale, loihi, large_scale_neuromorphic, review_of_neuromorphic}. Such hardware only consumes a fraction of the power required by DNNs on von Neumann computers and thus represents a suitable solution for energy-efficient edge computing \citep{benchmarking_keyword_spotting_neuromorphic, event_driven_visiual_tactile_sensing}. 
\par
The power consumption of neuromorphic hardware is closely related to the number of spikes they produce. Therefore, sparse SNNs that only fire a small number of spikes achieve high energy efficiency on hardware. However, such networks also transmit less information, creating a trade-off between energy consumption and model accuracy. In addition to sparsity, various tradeoffs between performance and other aspects of SNNs are commonly explored in the literature \citep{noise_robust_deep_snn, performance_sparsity_tradeoff, rethinking_gradiet_descent_for_snn, fast_classifying_high_accuracy_spiking_deep_networks}.
\par
For instance, unsupervised learning rules such as Spike Time Dependent Plasticity (STDP) can be implemented directly on neuromorphic hardware, allowing for biologically plausible and energy-efficient training of SNNs that are resilient to substrate noise of analog circuits \citep{stdp_hardware}. However, the performance of unsupervised learning lags behind that achieved with supervised learning. Meanwhile, state-of-the-art performance with SNNs is currently achieved through various error backpropagation (BP) techniques adapted from deep learning \citep{spikeprop, training_deep_snn_using_bp, macro_micro_backpropagation, S4NN, BS4NN, STDBP, slayer, spatio_temporal_backpropagation_snn, supervised_learning_based_on_temporal_coding, gradient_descent_alpha_function, deep_fast, incorporating_learnable_mp}. However, BP algorithms often require constraints on spikes to achieve high sparsity and low prediction latency, at the cost of performance \citep{bp_with_sparsity_regularization, supervised_learning_based_on_temporal_coding, snn_pruning, S4NN, BS4NN, deep_fast}. In addition, BP requires global transport of information that is incompatible with neuromorphic hardware and training must be performed either offline or in-the-loop, where a conventional computer is used in conjunction with neuromorphic hardware \citep{training_in_the_loop_brainscale, deep_fast}. Therefore, SNNs trained offline or in the loop must also be resilient to the substrate noise and weight quantization of analog hardware to avoid performance drops at deployment.
\par
One particularly interesting approach for training fast, energy-efficient and noise-resilient SNNs is Fast \& Deep \citep{deep_fast}. This exact BP method employs Time-To-First-Spike (TTFS) coding, which restricts neurons to fire only once. Inspired by the human visual system \citep{speed_of_processing_of_human_visual_system}, TTFS is based on the idea that first spikes of neurons must carry most of the information about input stimuli, enabling fast, sparse, and energy-efficient computation. Due to this constraint imposed on firing, we thus referred to TTFS networks as \textit{constrained} SNNs. However, relaxing the spike constraint of TTFS and allowing multiple spikes per neuron typically results in higher information rates, better performance, and increased noise resilience compared to TTFS networks \citep{macro_micro_backpropagation, spike_train_level_backpropagation, slayer, training_deep_snn_using_bp, STDBP}. Thus referred to as \textit{unconstrained} SNNs, one might assume that such network would improve performance and noise robustness but also result in slower inference and lower energy efficiency due to increased firing rates.
\par
In this work, we explore the tradeoffs between performance, convergence, energy consumption, prediction speed and robustness of SNNs, with and without the spike constraint imposed by TTFS. Our main contributions are:
\begin{itemize}
	\item We demonstrate that many aspects are driven by the weight distribution in Fast \& Deep, highlighting tradeoffs between performance, energy consumption, latency, and stability in TTFS SNNs.
	\item We extend the Fast \& Deep algorithm to multiple spikes per neuron and describe how errors are backpropagated in unconstrained SNNs.
	\item We show that our proposed method improves performance while providing better convergence rate, similar sparsity, comparable latency and improved robustness to noise compared to Fast \& Deep, suggesting that relaxing the spike constraints in TTFS can lead to better tradeoffs.
\end{itemize}
The rest of the article is structured as follows. In Section \ref{section:method}, we derive the closed-form solution for spike timings as well as their gradients, as described by Fast \& Deep \citep{deep_fast}, and describe how errors are backpropagated in unconstrained SNNs. In Section \ref{section:results} we show the results of our experiments on performance, convergence, sparsity, prediction latency, and robustness to noise and weight quantization of Fast \& Deep and our proposed method.

\section{Method} \label{section:method}

In this section, we describe our generalization of the Fast \& Deep algorithm to multiple spikes per neuron. Our main contribution lies in the reset of the membrane potential and how errors are backpropagated through inter-neuron and intra-neuron dependencies.

\subsection{The CuBa LIF Neuron}

We consider a neural network of Current-Based Leaky Integrate-and-Fire neurons with a soft reset of the membrane potential. \citep{spiking_neuron_models_single_neuron_populations_plasticity, loihi, deep_fast}.
\par
Formally, the dynamic of the membrane potential $u^{(l,j)}$ of the $j^{\text{th}}$ neuron in layer $l$ is given by the system of linear ordinary differential equations:
\begin{equation} \label{eq:lif_differential}
	\begin{split}
		\frac{du^{(l,j)}}{dt} &= -\frac{1}{\tau} u^{(l,j)}(t) + g^{(l,j)}(t) - \underbrace{\vartheta \delta\left(u^{(l,j)}(t) - \vartheta\right)}_{\textstyle{\text{Reset}}} \\
		\frac{dg^{(l,j)}}{dt} &= -\frac{1}{\tau_s}g^{(l,j)}(t) + \underbrace{\sum_{i=1}^{N^{(l-1)}} w_{i,j}^{(l)}  \sum_{z=1}^{n^{(l-1,i)}} \delta(t - t_z^{(l-1,i)})}_{\textstyle{\text{Pre-synaptic spikes}}}
	\end{split}
\end{equation}
where 
\begin{equation}
	\delta (x) = \begin{cases}
		+\infty & \text{ if } x=0 \\ 
		0 & \text{ otherwise }
	\end{cases}
\end{equation}
is the Dirac delta function that satisfies the identiy $\int_{-\infty}^{+\infty}\delta(x) dx = 1$.
We denote the number of neurons in the layer $l$ as $N^{(l)}$, the number of spikes fired by the neuron $j$ of this layer as $n^{(l,j)}$ and the $k^{\text{th}}$ spike of this neuron as $t_k^{(l,j)}$. 
Each pre-synaptic spike received at a synapse $i$ of a neuron $j$ induces an increase in post-synaptic current $g^{(l,j)}$ by an amount $w_{i,j}^{(l)}$, which defines the strength of the synaptic connection.
The post-synaptic current $g^{(l,j)}$ is then integrated into the membrane potential $u^{(l,j)}$. We denote by $\tau$ and $\tau_s$ the membrane and synaptic time constants that control the decay of the membrane potential and the post-synaptic current respectively.
Finally, when the membrane potential reaches the threshold $\vartheta$ the neuron emits a post-synaptic spike at time $t_k^{(l,j)}$ where $k$ is the index of the emitted spike and $u^{(l,j)}$ is reset to zero due to an instantaneous negative current of the size of the threshold.
\par
The reset of the membrane potential in Equation \ref{eq:lif_differential} is the major difference with the TTFS model used in Fast \& Deep \citep{deep_fast}. By resetting the membrane potential after post-synaptic spikes, our model allows for further integration of inputs and thus the firing of several spikes. Therefore, this relaxes the constraint on spike counts imposed by Fast \& Deep.

\subsection{SRM Mapping}

The Spike Response Model (SRM) is a generalization of the LIF neuron where the sub-threshold behavior of the neuron is defined by an integral over the past \citep{spiking_neuron_models_single_neuron_populations_plasticity, deep_fast, event_based_exact_gradient_snn}. This form is more convenient to derive as it is an explicit function of time. Formally, the SRM neuron defines the membrane potential as:
\begin{equation} \label{eq:SRM}
	\begin{split}
		u^{(l,j)} \left(t\right) =& \sum_{i=1}^{N^{(l-1)}} w_{i,j}^{(l)} \sum_{z=1}^{n^{(l-1,i)}} \epsilon \left(t - t_{z}^{(l-1,i)}\right) - \sum_{z=1}^{n^{(l,j)}} \eta \left(t - t_z^{(l,j)}\right)
	\end{split}
\end{equation}	
where $\epsilon(t)$ is the Post-Synaptic Potential (PSP) kernel that represents the response of the neuron to a pre-synaptic spike and $\eta(t)$ is the refractory kernel that defines the reset behavior. In Fast \& Deep \citep{deep_fast}, the refractory kernel $\eta(t) = 0$ is zero as no reset of the membrane potential is required.
\par
Using the definition of the CuBa LIF (Equation \ref{eq:lif_differential}), we find by integration the following PSP and refractory kernels \citep{spiking_neuron_models_single_neuron_populations_plasticity}:
\begin{equation}
	\begin{split}
		\epsilon(t) =& \Theta (t)\frac{\tau \tau_s}{\tau - \tau_s} \left[ \exp \left(\frac{-t}{\tau}\right) - \exp \left(\frac{-t}{\tau_s}\right) \right] \\
		\eta(t) =& \Theta (t)\vartheta \exp \left(\frac{-t}{\tau}\right)
	\end{split}
\end{equation}
where
\begin{equation}
	\Theta (x) := \begin{cases}
		1 & \text{ if } x > 0 \\ 
		0 & \text{ otherwise }
	\end{cases}
\end{equation}
is the Heavyside step function.

\subsection{Closed-Form Solution of Spike Timing}

Let us consider the $k^{\text{th}}$ spike timing $t_k^{(l,j)}$ of the neuron $j$ in layer $l$.
Fast \& Deep \citep{deep_fast} described that, by constraining the membrane time constant as twice the synaptic time constant ($\tau=2\tau_s$), the SRM mapping of the LIF neuron can be written as a polynomial of degree 2, such as:
\begin{equation} \label{eq:quadratic_form}
	0 = -a_k^{(l,j)} \exp\left(\frac{-t_k^{(l,j)}}{\tau}\right) ^2 + b_k^{(l,j)} \exp\left(\frac{-t_k^{(l,j)}}{\tau}\right) - c_k^{(l,j)}
\end{equation}
where $a_k^{(l,j)}$, $b_k^{(l,j)}$ and $c_k^{(l,j)}$ are three coefficients that depend on the definition of the model.
Equation \ref{eq:quadratic_form} can therefore be solved for $t_k^{(l,j)}$ by using the quadratic formula:
\begin{equation} \label{eq:spike_time_equation}
	t_k^{(l,j)} = \tau \ln \left[\frac{2a_k^{(l,j)}}{b_k^{(l,j)} + x_k^{(l,j)}}\right]
\end{equation}
with $x_k^{(l,j)} = \sqrt{\left(b_k^{(l,j)}\right)^2 - 4a_k^{(l,j)}c_k^{(l,j)}}$.
\par
This equation can be thus used to infer the spike trains of neurons in an event-based manner.
\par
According to the definition of our model, we find by factorizing Equation \ref{eq:SRM} into Equation \ref{eq:quadratic_form} the following coefficients:
\begin{equation}
	\begin{split}
		a_k^{(l,j)} :=& \sum_{i=1}^{N^{(l-1)}} w_{i,j}^{(l)} \sum_{z=1}^{n^{(l-1,i)}} \Theta \left (t_k^{(l,j)} - t_z^{(l-1,i)} \right ) \exp \left(\frac{t_z^{(l-1,i)}}{\tau_s}\right)
	\end{split}
\end{equation}
\begin{equation}
	\begin{split}
		b_k^{(l,j)} :=& \sum_{i=1}^{N^{(l-1)}} w_{i,j}^{(l)} \sum_{z=1}^{n^{(l-1,i)}} \Theta \left (t_k^{(l,j)} - t_z^{(l-1,i)} \right ) \exp \left(\frac{t_z^{(l-1,i)}}{\tau}\right) \\
		&- \frac{\vartheta}{\tau} \sum_{z=1}^{n^{(l,j)}} \Theta \left (t_k^{(l,j)} - t_z^{(l,j)} \right ) \exp \left(\frac{t_z^{(l,j)}}{\tau}\right) \\
	\end{split}
\end{equation}
and
\begin{equation}
	\begin{split}
		c :=& \; c_k^{(l,j)} = \frac{\vartheta}{\tau}
	\end{split}
\end{equation}
Note that the value of $c_k^{(l,j)}$ is common to every spike emitted by the neuron. We thus denote its value as $c$ for short. Moreover, only the definition of the coefficient $b_k^{(l,j)}$ differs from Fast \& Deep \citep{deep_fast} due to the reset of the membrane potential. For comparison, $b_k^{(l,j)}$ is defined as follows in Fast \& Deep \citep{deep_fast}:
\begin{equation}
	\begin{split}
		b_k^{(l,j)} :=& \sum_{i=1}^{N^{(l-1)}} w_{i,j}^{(l)} \sum_{z=1}^{n^{(l-1,i)}} \Theta \left (t_k^{(l,j)} - t_z^{(l-1,i)} \right) \exp \left(\frac{t_z^{(l-1,i)}}{\tau}\right)
	\end{split}
\end{equation}

\subsection{Spike Count Loss Function}

For each neuron $j$ in the output layer $o$, we define a spike count target $y_j$. The aim is to minimize the distance between the actual output spike counts and their corresponding targets. Therefore, we define the loss function as:
\begin{equation}
	\mathcal{L} := \frac{1}{2} \sum_{j=1}^{N^{(o)}} \left(y_j - n^{(o,j)}\right)^2
\end{equation}
where $o$ is the index of the output layer and $y_j$ is the spike count target associated with the same neuron.

\subsection{Gradient of Unconstrained Neurons}

Because the spike timing now has a closed-form solution, it becomes differentiable which allows the computation of an exact gradient. We first state the total change of weight between two neurons.
\par
Let $\delta_k^{(l,j)}$ be the error received by the spike $k$ of the neuron $j$ in layer $l$, the change of weight $\Delta w_{i,j}^{(l)}$ between the pre-synaptic neuron $i$ and the post-synaptic neuron $j$ of the layer $l$ is defined as a sum of all errors applied to their corresponding spike derivatives:
\begin{equation}
	\begin{split}
		\Delta w_{i,j}^{(l)} =& \sum_{k=1}^{n^{(l,j)}} \frac{\partial \mathcal{L}}{\partial t_k^{(l,j)}} \frac{\partial t_k^{(l,j)}}{\partial w_{i,j}^{(l)}} \\
		=& \sum_{k=1}^{n^{(l,j)}} \delta_k^{(l,j)} \frac{\partial t_k^{(l,j)}}{\partial w_{i,j}^{(l)}}
	\end{split}
\end{equation}
where
\begin{equation}
	\begin{split}
		\frac{\partial t_k^{(l,j)}}{\partial w_{i,j}^{(l)}} =& \sum_{z=1}^{n^{(l-1,i)}} \Theta \left (t_k^{(l,j)} - t_z^{(l-1,i)} \right ) \left[f_k^{(l,j)} \exp\left(\frac{t_z^{(l-1,i)}}{\tau_s}\right) - h_k^{(l,j)} \exp\left(\frac{t_z^{(l-1,i)}}{\tau}\right)\right]
	\end{split}
\end{equation}
is the partial derivative of Equation \ref{eq:spike_time_equation} with respect to the weight $w_{i,j}^{(l)}$ and:
\begin{equation}
	\begin{split}
		f_k^{(l,j)} :=& \frac{\partial t_k^{(l,j)}}{\partial a_k^{(l,j)}} = \frac{\tau}{a_k^{(l,j)}} \left[ 1 + \frac{c}{x_k^{(l,j)}} \exp \left( \frac{t_k^{(l,j)}}{\tau}\right)\right] \\
		h_k^{(l,j)} :=& \frac{\partial t_k^{(l,j)}}{\partial b_k^{(l,j)}} = \frac{\tau}{x_k^{(l,j)}}
	\end{split}
\end{equation}
\par
Therefore, the weight $w_{i,j}^{(l)}$ can be updated using the gradient descent algorithm as:
\begin{equation}
	w_{i,j}^{(l)} = w_{i,j}^{(l)} - \lambda \Delta w_{i,j}^{(l)}
\end{equation}
where $\lambda$ is a learning rate parameter.

\subsection{Spike Errors}

\begin{figure}[t]
	\hfill
	\begin{center}
		\includegraphics[width=0.7\columnwidth]{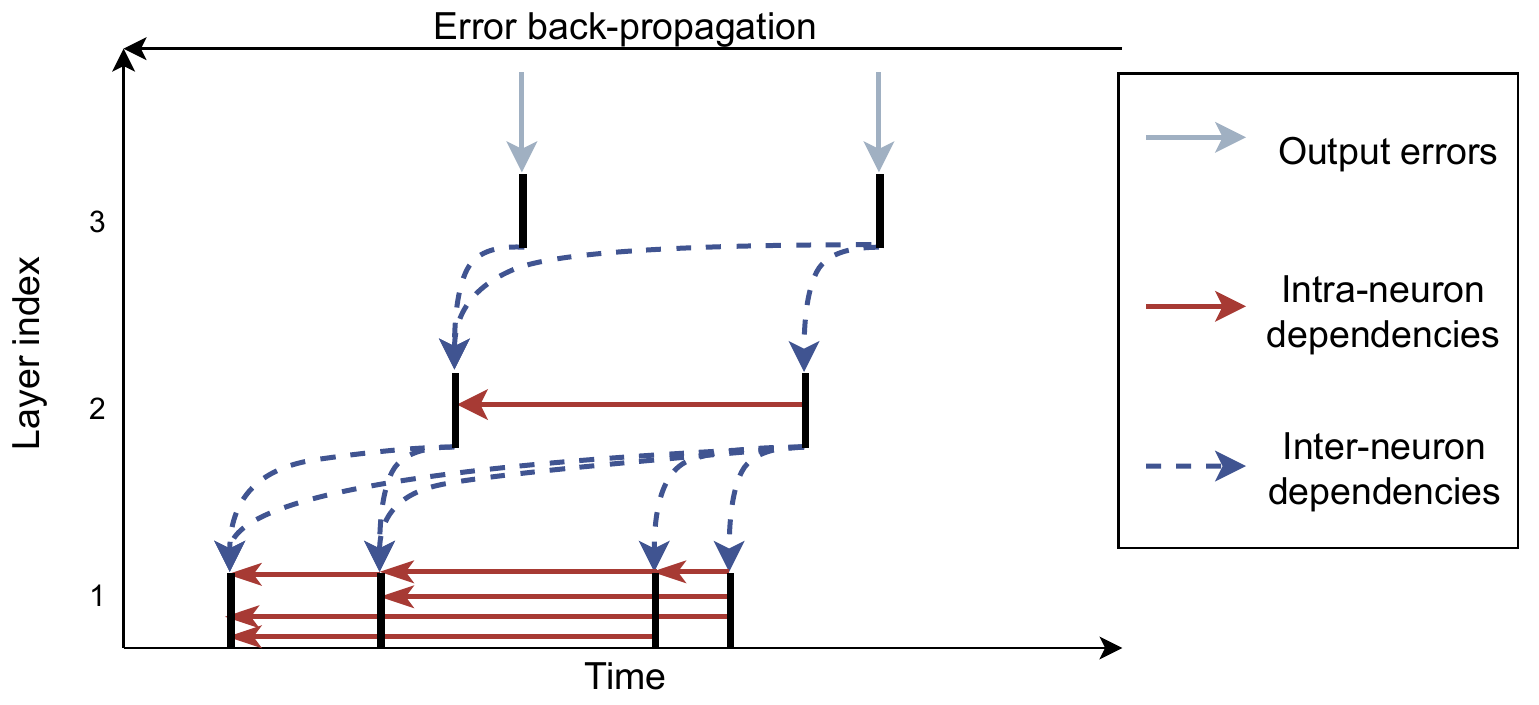}
	\end{center}
	\caption{Illustration of error backpropagation through spikes. This figure represents a three-layered network where the spike trains of only one neuron per layer are shown. The grey arrows represent the error coming from the loss function, the dashed blue arrows are the errors backpropagated from the downstream spikes (i.e. inter-neuron dependencies) and the red arrows are the error backpropagated from the future activity of the neuron due to the recurrence of the reset function (i.e. intra-neuron dependencies).}
	\label{fig:bats_soft_reset}
	\vspace{0.2in}
\end{figure}

We now derive the spike error $\delta_k^{(l,j)}$ associated to the spike at time $t_k^{(l,j)}$. In unconstrained SNNs, errors are backpropagated through two distincts paths: from post-synaptic to pre-synaptic spikes (i.e. inter-neuron dependencies) due to the synaptic connections and through post-synaptic to post-synaptic spikes (intra-neuron dependencies) due to the reset of the membrane potential after firing. In contrast, Fast \& Deep \citep{deep_fast} only propagates errors through inter-neuron dependencies as the membrane potential of constrained neurons is never reset. Figure \ref{fig:bats_soft_reset} provides a visual illustration of these dependencies. Therefore, we decompose the error received by the spike $t_k^{(l,j)}$ into two error components, such as:
\begin{equation}
	\begin{split}
		\delta_k^{(l,j)} :=& \frac{\partial \mathcal{L}}{\partial t_k^{(l,j)}} \\
		=& \underbrace{\sum_{i=1}^{N^{(l+1)}} \sum_{z=1}^{n^{(l+1,i)}} \frac{\partial \mathcal{L}}{\partial t_z^{(l+1,i)}} \frac{\partial t_z^{(l+1,i)}}{\partial t_k^{(l,j)}}}_{\textstyle{\text{Inter-Neuron}}} +  \underbrace{\sum_{z=k+1}^{n^{(l,j)}} \frac{\partial \mathcal{L}}{\partial t_z^{(l,j)}} \frac{\partial t_z^{(l,j)}}{\partial t_k^{(l,j)}}}_{\textstyle{\text{Intra-Neuron}}} \\
		=& \phi_k^{((l,j)} + \mu_k^{(l,j)}
	\end{split}
\end{equation}
where $\phi_k^{((l,j)}$ is the error backpropagated from inter-neuron dependencies and $\mu_k^{(l,j)}$ is the error backpropagated from intra-neuron dependencies.
\par
The inter-neuron error $\phi_k^{((l,j)}$ has different definitions for output neurons and hidden neurons. For output neurons, the inter-neuron error $\phi_k^{((o,j)}$ received by the $k^{\text{th}}$ spike of the neuron $j$ is defined as the derivative of the spike count loss function as no inter-neuron backpropagation is required:
\begin{equation} \label{eq:output_phi}
	\begin{split}
		\phi_k^{(o,j)} := \; \frac{\partial \mathcal{L}}{\partial n^{(o,j)}} = \; y_j - n^{(o,j)}
	\end{split}
\end{equation}
For hidden neurons, the inter-neuron error $\phi_k^{(l,j)}$ is defined as the sum of all errors backpropagated from the downstream spikes that have been emitted since $t_k^{(l,j)}$, such as:
\begin{equation}
	\begin{split}
		\phi_k^{(l,j)} :=& \sum_{i=1}^{N^{(l+1)}} \sum_{z=1}^{n^{(l+1,i)}} \frac{\partial \mathcal{L}}{\partial t_z^{(l+1,i)}} \frac{\partial t_z^{(l+1,i)}}{\partial t_k^{(l,j)}} \\
		=& \sum_{i=1}^{N^{(l+1)}} \sum_{z=1}^{n^{(l+1,i)}} \delta_z^{(l+1,i)} \frac{\partial t_z^{(l+1,i)}}{\partial t_k^{(l,j)}}
	\end{split}
\end{equation}
where
\begin{equation}
	\begin{split}
		\frac{\partial t_z^{(l+1,i)}}{\partial t_k^{(l,j)}} =& \Theta \left (t_z^{(l+1,i)} - t_k^{(l,j)} \right ) w_{j,i}^{(l+1)} \left[\frac{f_z^{(l+1,i)}}{\tau_s} \exp \left( \frac{t_k^{(l,j)}}{\tau_s} \right) - \frac{h_z^{(l+1,i)}}{\tau} \exp \left( \frac{t_k^{(l,j)}}{\tau} \right)\right]
	\end{split}
\end{equation}
For the intra-neuron error $\mu_k^{(l,j)}$, all the errors backpropagated from the future spike activity of the neuron must be taken into account due to the temporal impact of the resets on the future membrane potential. Therefore, $\mu_k^{(l,j)}$ is defined as a sum over all errors backpropagated from the following post-synaptic spikes in time:
\begin{equation} \label{eq:quadratic_intra_error}
	\begin{split}
		\mu_k^{(l,j)} :=& \sum_{z=k+1}^{n^{(l,j)}} \frac{\partial \mathcal{L}}{\partial t_z^{(l,j)}} \frac{\partial t_z^{(l,j)}}{\partial t_k^{(l,j)}} \\
		=& \sum_{z=k+1}^{n^{(l,j)}} \delta_z^{(l,j)} \frac{\partial t_z^{(l,j)}}{\partial t_k^{(l,j)}}
	\end{split}
\end{equation}
where 
\begin{equation}
	\frac{\partial t_z^{(l,j)}}{\partial t_k^{(l, j)}} =\frac{\vartheta}{\tau x_z^{(l,j)}} \exp \left(\frac{t_k^{(l,j)}}{\tau}\right)
\end{equation}
If evaluated as written, Equation \ref{eq:quadratic_intra_error} implies a quadratic time complexity $\mathcal{O}\left(n^2\right)$ (where $n$ is the total number of spikes emitted by the neurons) due to the recurrence. However, every reset has the same impact on the membrane potential i.e. an instantaneous negative current of the size of the threshold. Therefore, Equation \ref{eq:quadratic_intra_error} can be factorized as follows:
\begin{equation}
	\begin{split}
		\mu_k^{(l,j)} =& \sum_{z=k+1}^{n^{(l,j)}} \frac{\vartheta}{\tau x_z^{(l,j)}} \exp \left(\frac{t_k^{(l,j)}}{\tau}\right) \delta_z^{(l,j)} \\
		=& \frac{\vartheta}{\tau}\exp \left(\frac{t_k^{(l,j)}}{\tau}\right) \sum_{z=k+1}^{n^{(l,j)}} \frac{\delta_z^{(l,j)}}{x_z^{(l,j)}} \\
		=& \alpha_k^{(l,j)} \beta_k^{(l,j)}
	\end{split}
\end{equation}
where
\begin{equation}
	\alpha_k^{(l,j)} := \frac{\vartheta}{\tau} \exp \left(\frac{t_k^{(l,j)}}{\tau}\right)
\end{equation}
is the unique factor and
\begin{equation}
	\beta_k^{(l,j)} := \sum_{z=k+1}^{n^{(l,j)}} \frac{\delta_z^{(l,j)}}{x_z^{(l,j)}}
\end{equation}
is the backpropagated factor that can be accumulated in linear time complexity $\mathcal{O}\left(n\right)$ during the backward pass which significantly reduces the processing time.

\section{Results} \label{section:results}

In this section, we compare our proposed method with Fast \& Deep. This evaluation was conducted based on multiple criteria including performance on benchmark datasets, convergence rate, sparsity, classification latency, as well as robustness to noise and weight quantization.
\par
Experimental conditions were standardized for both methods, except for weight distributions and thresholds. Two uniform weight distributions ($w_{i,j} \sim U(-1, 1)$ and $w_{i,j} \sim U(0, 1)$) were used to evaluate Fast \& Deep, to measure the effect of initial weight distributions on the different evaluation criteria. Our method was solely assessed using $w_{i,j} \sim U(-1, 1)$, as positive initial weights lead to excessive spiking activity, hindering computational and energy efficiency. Thresholds were manually tuned to find the best-performing networks and kept fixed during training. In our experiments, all layers (including convolutional layers) have been directly trained using our proposed method and Fast \& Deep and no conversion from DNN to SNN has been performed. More details about our experimental settings can be found in Appendix.

\subsection{Performance and Convergence Rate}

\begin{table}[!tbp]
	\caption{Performances comparison between Fast \& Deep and our method on the MNIST, EMNIST, Fashion-MNIST and Spiking Heidelberg Digits (SHD) datasets. The initial weight distribution used in each column is specified in the first row of the table. Conv. refers to a convolutional SNN with the following architecture: 15C5-P2-40C5-P2-300-10.}
	\label{table:performance_comparison}
	\begin{center}
		\begin{tabular}{|c|cccc|}
			\hline
			\multirow{2}{*}{Dataset} & \multirow{2}{*}{Architecture} & \multirow{2}{*}{\begin{tabular}[c]{@{}l@{}}Fast \& Deep\\ $U(-1, 1)$ \end{tabular}} & \multirow{2}{*}{\begin{tabular}[c]{@{}l@{}}Fast \& Deep\\ $U(0, 1)$ \end{tabular}} & \multirow{2}{*}{\begin{tabular}[c]{@{}l@{}}Our Method\\ $U(-1, 1)$ \end{tabular}} \\
			&  &  &  & \\ \hline
			\multirow{2}{*}{MNIST} & 800-10 & 96.76 $\pm$ 0.17\% & 97.83 $\pm$ 0.08\% & \textbf{98.88 $\boldsymbol{\pm}$ 0.02\%} \\
			& Conv. & 99.01 $\pm$ 0.16\% & 99.22 $\pm$ 0.05\% & \textbf{99.38 $\boldsymbol{\pm}$ 0.04\%} \\ \hline
			EMNIST & 800-47 & 69.56 $\pm$ 6.70\% & 83.34 $\pm$ 0.27\% & \textbf{85.75 $\boldsymbol{\pm}$ 0.06\%} \\ \hline
			Fashion MNIST & 400-400-10 & 88.14 $\pm$ 0.08\% & 88.47 $\pm$ 0.20\% & \textbf{90.19 $\boldsymbol{\pm}$ 0.12\%} \\ \hline
			SHD & 128-20 & 33.84 $\pm$ 1.35\% & 47.37 $\pm$ 1.65\% & \textbf{66.8 $\boldsymbol{\pm}$ 0.76\%} \\ \hline
		\end{tabular}
	\end{center}
\end{table}

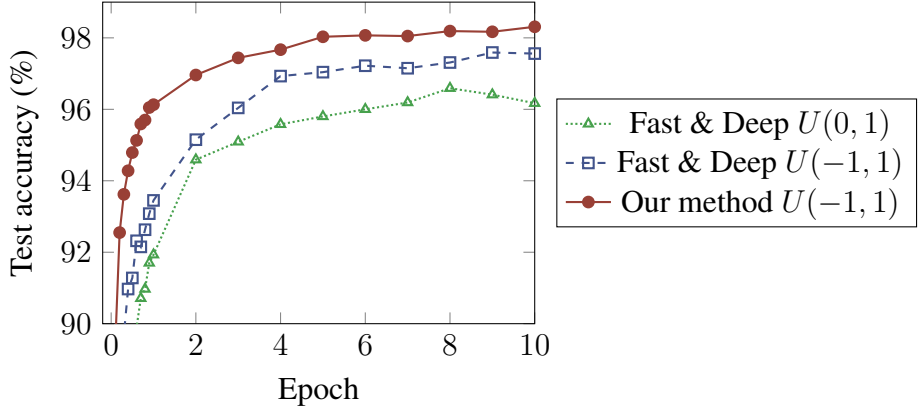
\begin{figure}[!tbp]
	\centering
	\begin{tikzpicture}[trim axis left, trim axis right]
		\begin{axis}[
			xmin=-0.2, xmax=10,
			ymin=90, ymax=99,
			xlabel={Epoch},
			ylabel={Test accuracy (\%)},
			legend style={at={(1.05, 0.5)}, anchor=west},
			width=0.5\linewidth,
			height=0.4\linewidth]
			
			\addplot[thick, color=greenmat, mark=triangle, mark options={solid}, densely dotted] table [x=Epoch, y=Accuracy TTFS -1 1, col sep=comma] {Figures/test_acc.csv};
			\addplot[thick, color=bluemat, mark=square, mark options={solid}, dashed] table [x=Epoch, y=Accuracy TTFS 0 1, col sep=comma] {Figures/test_acc.csv};
			\addplot[thick, mark=*, mark options={solid}, color=redmat] table [x=Epoch, y=Accuracy Rate, col sep=comma] {Figures/test_acc.csv};
			\legend{{Fast \& Deep $U(0, 1)$}, {Fast \& Deep $U(-1, 1)$}, {Our method $U(-1, 1)$}}
		\end{axis}
	\end{tikzpicture}
	\caption{Test accuracy (Figure \ref{fig:test_accuracy_mnist}) of Fast \& Deep and our method on MNIST for the same learning rate. The unconstrained SNN trained with our method benefits from a higher convergence rate than the temporaly-coded networks trained with Fast \& Deep. The two SNNs trained with Fast \& Deep have similar convergence rates despite their difference in initial weight distribution.}
	\label{fig:test_accuracy_mnist}
\end{figure}

\begin{figure}[!tbp]
	\centering
	
	\begin{subfigure}[t]{0.5\textwidth}
		\centering
		\hspace{-1.0cm}
		\begin{tikzpicture}[trim axis right]
			\begin{axis}[
				xmin=0, xmax=1.0,
				ymin=0, ymax=750,
				ylabel={Neuron index},
				legend style={at={(1.05, 0.5)}, anchor=west},
				width=1.0\linewidth,
				height=0.7\linewidth]
				
				\addplot+[only marks, color=black, mark options={fill=black,scale=0.3}] table [x=Time, y=Index, col sep=comma] {Figures/inputs_spikes_shd.csv};
			\end{axis}
		\end{tikzpicture}
		\caption{Inputs}
		\label{fig:inputs_spikes_shd}
	\end{subfigure}%
		\begin{subfigure}[t]{0.5\textwidth}
		\centering
		\hspace{-1.0cm}
		\begin{tikzpicture}[trim axis right]
			\begin{axis}[
				xmin=0, xmax=1.0,
				ymin=0, ymax=128,
				legend style={at={(1.05, 0.5)}, anchor=west},
				width=1.0\linewidth,
				height=0.7\linewidth]
				
				\addplot+[only marks, color=black, mark options={fill=black,scale=0.3}] table [x=Time, y=Index, col sep=comma] {Figures/rate_spikes_shd.csv};
			\end{axis}
		\end{tikzpicture}
		\caption{Inputs}
		\label{fig:rate_spikes_shd}
	\end{subfigure}
	\begin{subfigure}[t]{0.5\textwidth}
		\centering
		\hspace{-1.0cm}
		\begin{tikzpicture}[trim axis right]
			\begin{axis}[
				xmin=0, xmax=1.0,
				ymin=0, ymax=128,
				xlabel={Time (s)},
				ylabel={Neuron index},
				legend style={at={(1.05, 0.5)}, anchor=west},
				width=1.0\linewidth,
				height=0.7\linewidth]
				
				\addplot+[only marks, color=black, mark options={fill=black,scale=0.3}] table [x=Time, y=Index, col sep=comma] {Figures/ttfs_1_1_spikes_shd.csv};
			\end{axis}
		\end{tikzpicture}
		\caption{Fast \& Deep $U(-1,1)$}
		\label{fig:ttfs_1_1_spikes_shd}
	\end{subfigure}%
	\begin{subfigure}[t]{0.5\textwidth}
		\centering
		\hspace{-1.0cm}
		\begin{tikzpicture}[trim axis right]
			\begin{axis}[
				xmin=0, xmax=1.0,
				ymin=0, ymax=128,
				xlabel={Time (s)},
				legend style={at={(1.05, 0.5)}, anchor=west},
				width=1.0\linewidth,
				height=0.7\linewidth]
				
				\addplot+[only marks, color=black, mark options={fill=black,scale=0.3}] table [x=Time, y=Index, col sep=comma] {Figures/ttfs_0_1_spikes_shd.csv};
			\end{axis}
		\end{tikzpicture}
		\caption{Fast \& Deep $U(0,1)$}
		\label{fig:ttfs_0_1_spikes_shd}
	\end{subfigure}%
	\caption{Spiking activity of hidden neurons in a SNN trained with our method (Figure \ref{fig:rate_spikes_shd}) and Fast \& Deep (Figures \ref{fig:ttfs_1_1_spikes_shd} and \ref{fig:ttfs_0_1_spikes_shd}) given a spoken "zero" (Figure \ref{fig:inputs_spikes_shd}) from the SHD dataset. TTFS neurons in Fast \& Deep manly respond to early stimuli, missing most of the input information. In contrast, our method allows for multiple spikes per neuron which enables them to capture all the information from the inputs. This demonstrates the importance of relaxing the spike constraint of TTFS when processing temporal data.}
	\label{fig:hidden_spikes_shd}
\end{figure}
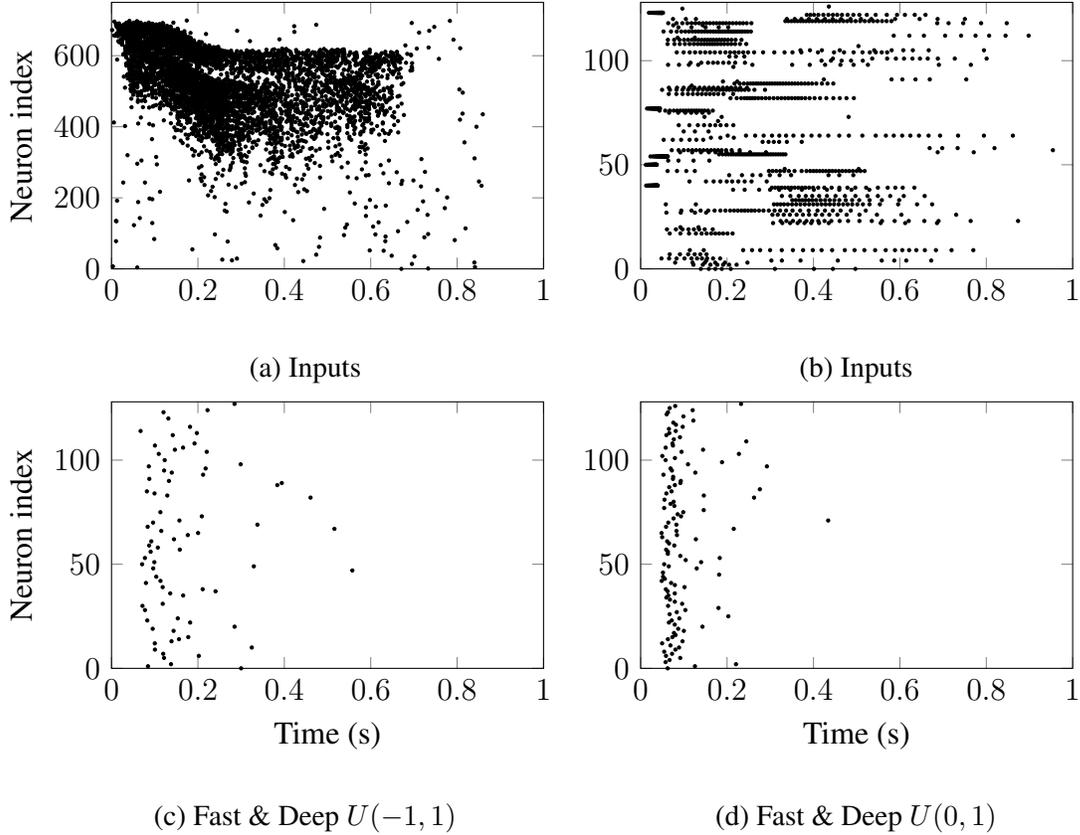

To assess the performance of our proposed method, we trained fully-connected SNNs on the MNIST \citep{mnist}, EMNIST \citep{emnist} (Balanced Extended MNIST) and Fashion-MNIST \citep{fashion_mnist} datasets as well as Convolutional SNNs on MNIST. We also evaluated our method on temporal data classification by training fully-connected networks on the Spiking Heidelberg Digits (SHD) dataset \citep{heidelberg_dataset}, an english and german spoken digits classification task. We compared our results to those obtained using the original Fast \& Deep algorithm. Table \ref{table:performance_comparison} summarizes the average test accuracies of both methods given the considered initial weight distributions. For completeness, a comparison of Fast \& Deep and our method with other spike-based BP algorithms can be found in Appendix.
\par
Firstly, it should be noted that the Fast \& Deep algorithm generally achieves better performance when the weights are initialized with positive values, which is consistent with the choice of weight distribution made by \cite{deep_fast}. Secondly, our proposed method demonstrates superior performance compared to the Fast \& Deep algorithm, with improvement margins ranging from 1.05\% on MNIST to 2.41\% on the more difficult EMNIST dataset. This is not surprising given that unconstrained SNNs are known to perform better compared to TTFS networks \citep{macro_micro_backpropagation, spike_train_level_backpropagation, slayer, training_deep_snn_using_bp, STDBP}. Moreover, our methods outperforms by at least 19\% Fast \& Deep on the SHD dataset. To understand the reason for this performance gap, we analyzed the spiking activity in the hidden layers after training. Figure \ref{fig:hidden_spikes_shd} shows that TTFS neurons only respond to early stimuli. In temporal coding, high-valued information is encoded by early spikes. Training with a one spike constraint is therefore energy efficient but tends to make SNNs spike as early as possible, thus missing the information occuring later in time. In contrast, neurons trained without spike constraint are able to response throughout the duration of the sample, thus capturing all the information despite an increased total of spikes fired. This highlights the importance of firing more than once and demonstrate that a tradeoff exists between performance and energy consumption when processing temporal data.
\par
In addition, our results indicate that the convergence rate of SNNs with multiple spikes is higher compared to TTFS networks. Figure \ref{fig:test_accuracy_mnist} depicts the evolution of the test accuracy of both methods on MNIST. This demonstrates that our method can reach desired accuracies in fewer epochs compared to Fast \& Deep. Such improvement implies that discriminative features are learnt earlier during training.

\subsection{Network Sparsity}

\begin{figure}[!tbp]
	\centering
	\begin{tikzpicture}[trim axis left, trim axis right]
		\begin{axis} [ybar=10pt,
			xmin=0, xmax=4,
			ymin=0, ymax=800,
			xtick={1, 3},
			xticklabels={Spike count, Active neurons},
			nodes near coords,
			nodes near coords align={vertical},
			legend style={at={(1.05, 0.5)}, anchor=west}, 
			cells={align=center},
			width=0.6\linewidth,
			height=0.4\linewidth,
			]
			\addplot[fill=bluemat!20, postaction={
				pattern=north east lines,
				pattern color=bluemat
			}] coordinates {
				(1,660) 
				(3,660)
			};
			\addplot[fill=greenmat!20, postaction={
				pattern=north east lines,
				pattern color=greenmat
			}] coordinates {
				(1,192) 
				(3,192)
			};
			\addplot[fill=redmat!20, postaction={
				pattern=dots,
				pattern color=redmat
			}] coordinates {
				(1,216)
				(3,54)
			};
			\legend{{Fast \& Deep $U(0, 1)$}, {Fast \& Deep $U(-1, 1)$}, {Our method $U(-1, 1)$}}
		\end{axis}
	\end{tikzpicture}
	\caption{Comparison of the population spike count and the number of active neurons in fully-connected SNNs trained using Fast \& Deep and our proposed method on the MNIST dataset. These results indicate that the sparsity of SNNs after training depends on the weight distribution, with the SNN initialized with only positive weights appearing to be less sparse than those initialized with both negative and positive weights. Additionally, our proposed method demonstrates a similar level of sparsity and fewer active neurons as Fast \& Deep for the same initial weight distribution, despite the relaxed constraint on neuron spike counts.}
	\label{fig:spike_count_active}
	\vspace{0.2in}
\end{figure}
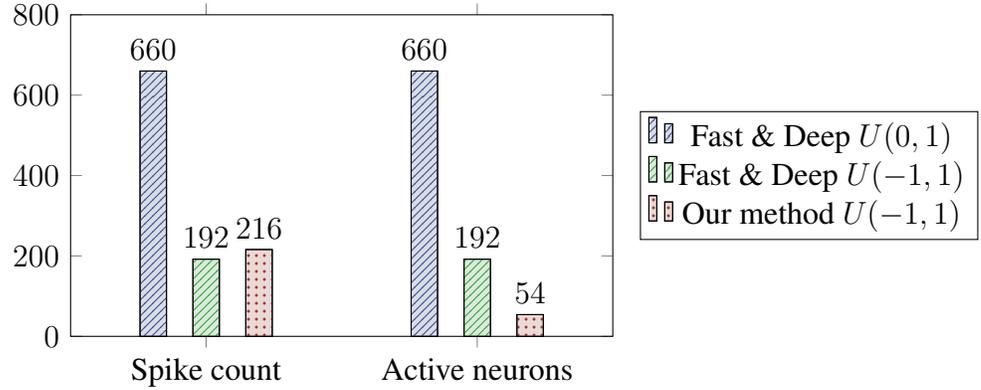

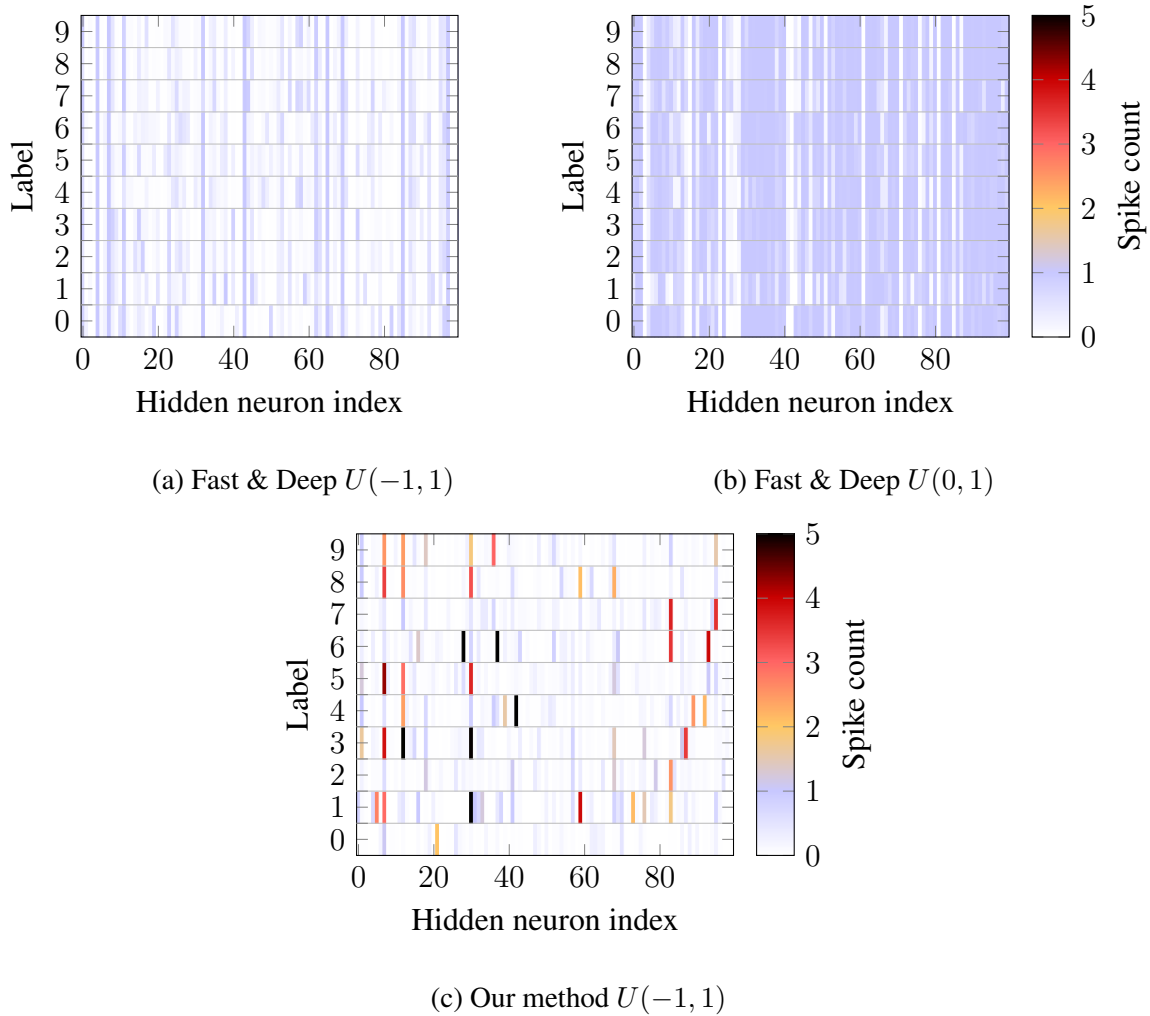
\begin{figure}[!tbp]
	\centering
	\begin{subfigure}[t]{0.5\textwidth}
		\centering
		\hspace{-1.0cm}
		\begin{tikzpicture}[trim axis left, trim axis right]
			\begin{axis}[
				axis on top,
				xmin=0,
				xmax=99,
				ymin=0,
				ymax=9,
				ytick={0,1,2,3,4,5,6,7,8,9},
				extra y ticks={0.5,1.5,2.5,3.5,4.5,5.5,6.5,7.5,8.5},
				extra y tick labels={},
				extra y tick style={
					grid=major,
				},
				enlargelimits={abs=0.5},
				point meta=explicit,
				colormap={my colormap}{
					rgb255=(255, 255, 255),
					rgb255=(200, 200, 255),
					rgb255=(255, 200, 100),
					rgb255=(255, 100, 100),
					rgb255=(200, 0, 0),
					rgb255=(0, 0, 0),
				},
				colorbar style={
					ylabel={Spike count},
					ytick={0, 1, 2, 3, 4, 5},
				},
				width=0.9\linewidth,
				height=0.8\linewidth,
				point meta min=0,
				point meta max=5,
				xlabel={Hidden neuron index},
				ylabel={Label}]
				
				\addplot [matrix plot*, mesh/cols=200, point meta=explicit] table [x=Neuron_index, y=Label, meta=TTFS 1 1, col sep=comma] {Figures/activity_distribution.csv};
			\end{axis}
		\end{tikzpicture}
		\caption{Fast \& Deep $U(-1,1)$}
		\label{fig:activity_fast_and_deep_11}
	\end{subfigure}%
	\begin{subfigure}[t]{0.5\textwidth}
		\centering
		\hspace{-1.0cm}
		\begin{tikzpicture}[trim axis left, trim axis right]
			\begin{axis}[
				axis on top,
				xmin=0,
				xmax=99,
				ymin=0,
				ymax=9,
				ytick={0,1,2,3,4,5,6,7,8,9},
				extra y ticks={0.5,1.5,2.5,3.5,4.5,5.5,6.5,7.5,8.5},
				extra y tick labels={},
				extra y tick style={
					grid=major,
				},
				enlargelimits={abs=0.5},
				point meta=explicit,
				colormap={my colormap}{
					rgb255=(255, 255, 255),
					rgb255=(200, 200, 255),
					rgb255=(255, 200, 100),
					rgb255=(255, 100, 100),
					rgb255=(200, 0, 0),
					rgb255=(0, 0, 0),
				},
				%colormap access=piecewise const,
				colorbar style={
					ylabel={Spike count},
					ytick={0, 1, 2, 3, 4, 5},
				},
				colorbar,
				width=0.9\linewidth,
				height=0.8\linewidth,
				point meta min=0,
				point meta max=5,
				xlabel={Hidden neuron index},
				ylabel={Label}]
				
				\addplot [matrix plot*, mesh/cols=200, point meta=explicit] table [x=Neuron_index, y=Label, meta=TTFS 0 1, col sep=comma] {Figures/activity_distribution.csv};
			\end{axis}
		\end{tikzpicture}
		\caption{Fast \& Deep $U(0,1)$}
		\label{fig:activity_fast_and_deep_01}
	\end{subfigure}
	\begin{subfigure}[t]{0.5\textwidth}
		\centering
		\hspace{-1.0cm}
		\begin{tikzpicture}[trim axis left, trim axis right]
			\begin{axis}[
				axis on top,
				xmin=0,
				xmax=99,
				ymin=0,
				ymax=9,
				ytick={0,1,2,3,4,5,6,7,8,9},
				extra y ticks={0.5,1.5,2.5,3.5,4.5,5.5,6.5,7.5,8.5},
				extra y tick labels={},
				extra y tick style={
					grid=major,
				},
				enlargelimits={abs=0.5},
				point meta=explicit,
				colormap={my colormap}{
					rgb255=(255, 255, 255),
					rgb255=(200, 200, 255),
					rgb255=(255, 200, 100),
					rgb255=(255, 100, 100),
					rgb255=(200, 0, 0),
					rgb255=(0, 0, 0),
				},
				%colormap access=piecewise const,
				colorbar style={
					ylabel={Spike count},
					ytick={0, 1, 2, 3, 4, 5},
				},
				colorbar,
				width=0.9\linewidth,
				height=0.8\linewidth,
				point meta min=0,
				point meta max=5,
				xlabel={Hidden neuron index},
				ylabel={Label}]
				
				\addplot [matrix plot*, mesh/cols=200, point meta=explicit] table [x=Neuron_index, y=Label, meta=Rate, col sep=comma] {Figures/activity_distribution.csv};
			\end{axis}
		\end{tikzpicture}
		\caption{Our method $U(-1,1)$}
		\label{fig:activity_bats}
	\end{subfigure}
	\caption{Figures \ref{fig:activity_fast_and_deep_11} and Figure \ref{fig:activity_fast_and_deep_01} show the average spike count of hidden neurons trained with Fast \& Deep on the MNIST dataset, while Figure \ref{fig:activity_bats} shows the average spike count of hidden neurons trained with our proposed method. Each row corresponds to the average activity over all the test samples of a particular digit. As TTFS networks mainly encode information temporally, we observe that neurons trained with Fast \& Deep fire indiscriminately in response to stimuli, making it difficult to differentiate the labels from the mean spike count, regardless of the initial weight distribution. However, our proposed SNN training method results in a different distribution of firing activity. More precisely, key neurons respond selectively to particular digits, while most of the other neurons remain mostly silent.}
	\label{fig:activity}
	\vspace{0.2in}
\end{figure}

\begin{figure}[!tbp]
	\centering
	
	\begin{subfigure}[t]{0.5\textwidth}
		\centering
		\begin{tikzpicture}[trim axis left]
			\begin{axis}
				[
				legend pos=north west,
				ymajorgrids=true,
				grid style=dotted,
				xlabel={Output Threshold},
				ylabel={Hidden spike count},
				width=0.9\linewidth,
				height=0.7\linewidth,
				]
				
				%\addplot[color=black, mark=o, dashed] coordinates {
				%	(0.4, 753)(0.7, 753)(1.0, 753)(1.3, 753)(1.6, 753)
				%};
				\addplot[color=redmat, mark=*] coordinates {
					(0.4, 247)(0.7, 409)(1.0,597)(1.3,745)(1.6, 906)
				};
				%\legend{Initial, 1 epoch};
			\end{axis}
		\end{tikzpicture}
	\caption{}
	\label{fig:hidden_count_vs_threshold}
	\end{subfigure}%
	\begin{subfigure}[t]{0.5\textwidth}
		\centering
		\begin{tikzpicture}[trim axis left]
			\begin{axis}
				[
				ymajorgrids=true,
				grid style=dotted,
				xlabel={Output Threshold},
				ylabel={Output spike count},
				width=0.9\linewidth,
				height=0.7\linewidth,
				]
				
				%\addplot[color=black, mark=square] coordinates {
				%	(0.4, 40.54) (0.7, 41.18) (1.0, 40.99) (1.3, 40.63)(1.6, 40.56)
				%};
				\addplot[color=redmat, mark=*] coordinates {
					(0.4, 18.55) (0.7, 8.98) (1.0, 5.27) (1.3, 3.37) (1.6, 2.24)
				};
				%\legend{Initial};
			\end{axis}
		\end{tikzpicture}
	\caption{}
	\label{fig:output_count_vs_threshold}
	\end{subfigure}
	\begin{subfigure}[t]{0.5\textwidth}
		\centering
		\begin{tikzpicture}[trim axis left]
			\begin{axis}
				[
				ymajorgrids=true,
				grid style=dotted,
				xlabel={Output Threshold},
				ylabel={Output errors},
				width=0.9\linewidth,
				height=0.7\linewidth,
				]
				
				\addplot[color=redmat, mark=*] coordinates {
					(0.4, -0.061746)(0.7, 0.039505836) (1.0, 0.08649886) (1.3, 0.114457116) (1.6, 0.1345539)
				};
				%\addplot[color=red, mark=*] coordinates {
				%	(0.2, 3.4821344e-06)(0.5, -9.028935e-07)(0.8, -1.0533839e-06)(0.9, -8.6623436e-07)(1.0, 0)(1.2, 0)
				%};
			\end{axis}
		\end{tikzpicture}
		\caption{}
		\label{fig:output_errors_vs_threshold}
	\end{subfigure}%
	\begin{subfigure}[t]{0.5\textwidth}
		\centering
		\begin{tikzpicture}[trim axis left]
			\begin{axis}
				[
				ymajorgrids=true,
				grid style=dotted,
				xlabel={Output Threshold},
				ylabel={Change of weights},
				width=0.9\linewidth,
				height=0.7\linewidth,
				]
				
				\addplot[color=redmat, mark=*] coordinates {
					(0.4, -0.00016658746)(0.7, -6.444694e-05) (1.0, -2.941604e-05) (1.3, -1.7438315e-05) (1.6, -9.303185e-06)
				};
				%\addplot[color=red, mark=*] coordinates {
				%	(0.2, 3.4821344e-06)(0.5, -9.028935e-07)(0.8, -1.0533839e-06)(0.9, -8.6623436e-07)(1.0, 0)(1.2, 0)
				%};
			\end{axis}
		\end{tikzpicture}
		\caption{}
		\label{fig:delta_w_vs_threshold}
	\end{subfigure}
	\caption{Influence of the output threshold on the sparsity of a 2-layer SNN trained on MNIST with our method. Figure \ref{fig:hidden_count_vs_threshold} illustrates that a lower output threshold results in fewer spikes generated after 1 epoch. Figure \ref{fig:output_count_vs_threshold} indicates that decreasing the output threshold increases the initial activity in the output layer, thereby leading to a greater number of negative errors transmitted during the backward pass (as shown in Figure \ref{fig:output_errors_vs_threshold}). This, in turn, leads to a decrease in the weights in the hidden layer, as depicted in Figure \ref{fig:delta_w_vs_threshold}.}
	\label{fig:cost_vs_accuracy}
\end{figure}
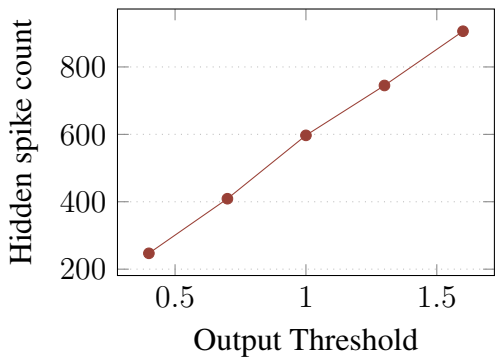
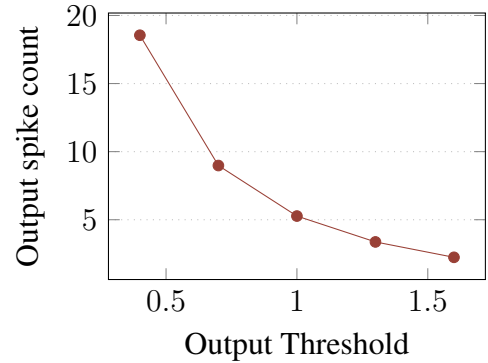
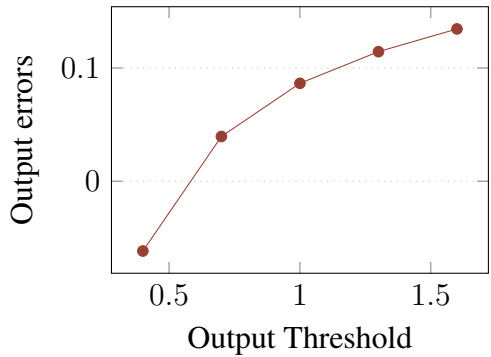
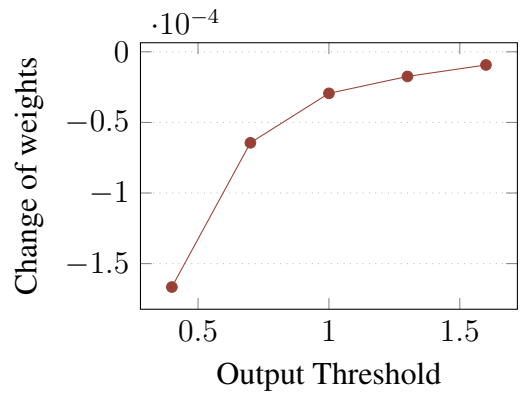

Achieving high sparsity in trained SNNs is critical for energy efficiency, as neuromorphic hardware only consumes energy at spike events.
\par
Figure \ref{fig:spike_count_active} shows the population spike counts of fully-connected SNNs trained on MNIST using both methods. It appears that the initial weight distribution plays an important role in the final sparsity level of the trained SNNs. In the analyzed case, SNNs initialized with positive weights appear to be less sparse after training than SNNs initialized with both negative and positive weights.
\par
While our proposed method allows for an increased number of spikes per neuron, which implies more energy consumption, we found that it can achieve similar sparsity as TTFS networks initialized with both negative and positive weights while performing better than TTFS networks initialized with positive weights only (see Figure \ref{fig:spike_count_active} and Table \ref{table:performance_comparison}). This suggests that our proposed method can offer improved tradeoffs between accuracy and sparsity.
\par
To understand why our method can achieve such levels of sparsity despite not imposing any constraints on neuron firing, we analyzed the average activity in each network. Figure \ref{fig:activity} shows that neurons trained with Fast \& Deep fire indiscriminately in response to any input digit, which is characteristic of temporal coding where information is represented by the timing of spikes, rather than the presence or absence of spikes. In contrast, SNNs trained with our method exhibit a different distribution of firing activity, with certain key neurons selectively responding to specific digits. Figure \ref{fig:spike_count_active} indicates that only 7\% of neurons trained with our method are active during inference (i.e. neurons that fire at least once). In comparison, the SNN trained with Fast \& Deep and initialized with both negative and positive weights has 24\% of its neurons firing. Therefore, the reduced proportion of active units in our method compensates for the increased number of spikes per neuron, leading to fewer spikes emitted in the network.
\par
We found during our experiments that the sparsity of SNNs trained with our method was also influenced by the choice of threshold values. More precisely, we observed that decreasing the output threshold resulted in a reduction of activity in the network, as illustrated in Figure \ref{fig:hidden_count_vs_threshold}. This decrease in activity occurred because the output neurons fired more often, which is illustrated in Figure \ref{fig:output_count_vs_threshold}. The increased number of output spikes caused the loss function to produce more negative errors, which resulted in negatives change of weights in the hidden layer. However, lowering the threshold has a dual impact on activity: it increases the firing rates at initialization but also contributes to producing more negative errors, which can decrease activity during learning. While controlling sparsity using thresholds is trivial in shallow SNNs, a fine balance between thresholds has to be found to control sparsity in Deep SNNs, making it challenging to achieve sparsity with larger architectures. However, our findings suggest that threshold values play a crucial role in determining the sparsity of unconstrained SNNs and can be seen as a way to control activity without the need for firing rate regularization.

\subsection{Prediction Latency}

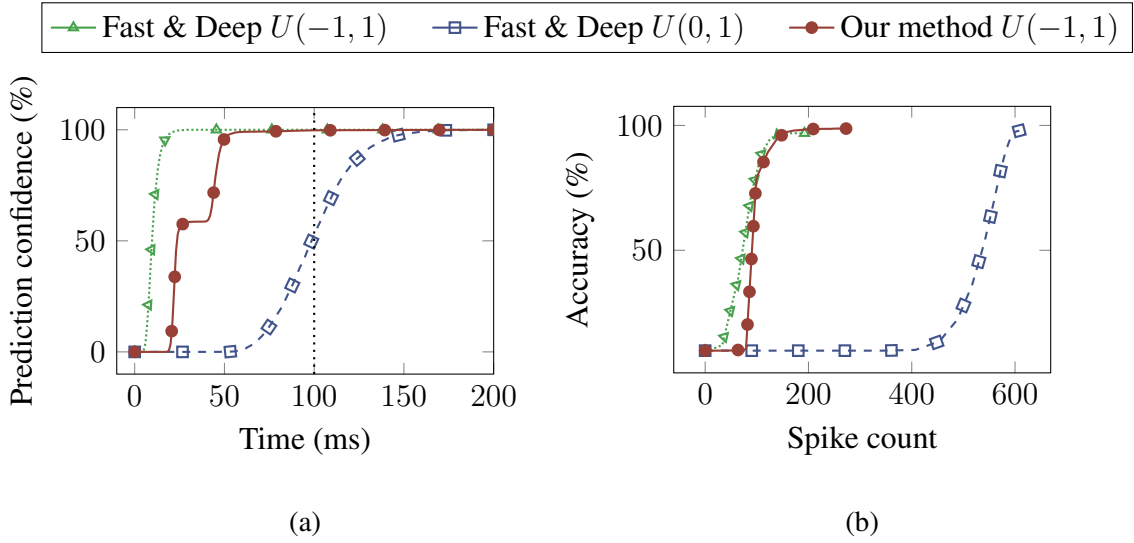
\begin{figure}[!tbp]
	\begin{subfigure}[t]{0.5\textwidth}
		\centering
		\begin{tikzpicture}[trim axis left, trim axis right]
			\begin{axis}[
				xmin=-10, xmax=200,
				ymin=-10, ymax=110,
				xlabel={Time (ms)},
				ylabel={Prediction confidence (\%)},
				legend style={
					at={(-0.2, 1.3)}, 		
					anchor=west,
					cells={align=center},
					/tikz/every even column/.append style={column sep=0.5cm},
					legend columns=-1
				},
				width=0.9\linewidth,height=0.7\linewidth]
				
				\addplot[densely dotted, greenmat, mymark={triangle}{solid}, thick] table [x=Time, y=TTFS -1 1, col sep=comma] {Figures/classification_latency.csv};
				\addplot[dashed, bluemat, mymark={square}{solid}, thick] table [x=Time, y=TTFS 0 1, col sep=comma] {Figures/classification_latency.csv};
				\addplot[redmat, mymark={*}{solid}, thick] table [x=Time, y=Rate, col sep=comma] {Figures/classification_latency.csv};
				\addplot +[mark=none, thick, dotted] coordinates {(100, -10) (100, 110)};
				\legend{{Fast \& Deep $U(-1, 1)$}, {Fast \& Deep $U(0, 1)$}, {Our method $U(-1, 1)$}}
			\end{axis}
		\end{tikzpicture}
		\caption{}
		\label{fig:prediction_confidence}
	\end{subfigure}
	\begin{subfigure}[t]{0.5\textwidth}
		\centering
		\begin{tikzpicture}[trim axis left, trim axis right]
			\begin{axis}
				[
				xlabel={Spike count},
				ylabel={Accuracy (\%)},
				width=0.9\linewidth,
				height=0.7\linewidth,
				]
				
				\addplot[thick, mymark={triangle}{solid}, color=greenmat, densely dotted] table[x=TTFS -1 1 Cost, y=TTFS -1 1 Acc, col sep=comma] {Figures/cost_vs_acc_mnist.csv};
				\addplot[thick, mymark={square}{solid}, color=bluemat, dashed] table[x=TTFS 0 1 Cost, y=TTFS 0 1 Acc, col sep=comma] {Figures/cost_vs_acc_mnist.csv};
				\addplot[thick, mymark={*}{solid}, color=redmat] table[x=Rate cost, y=Rate Acc, col sep=comma] {Figures/cost_vs_acc_mnist.csv};
			\end{axis}
		\end{tikzpicture}
		\caption{}
		\label{fig:spike_count_vs_accuracy}
	\end{subfigure}
	\caption{Figure \ref{fig:prediction_confidence} shows the evolution of the average prediction confidence during simulations on the MNIST test set. To produce this figure, we measured the probability of predictions at a time $t$ being equal to the final predictions at the end of the simulations. The vertical dotted line represents the end of input spikes. The initial weight distribution seems to have a crutial impact on the latency of predictions. More precisely, negative initial weights produce confidence earlier than positive initial weights. Therefore, simulation time can be reduced to further improve sparsity. Figure \ref{fig:spike_count_vs_accuracy} shows the relationship between spike count and accuracy as the simulation time increases. It demonstrates that the duration of simulation can be used as a post-training method to further reduce energy consumption while maintaining high performance.}
	\label{fig:mnist_latency}
\end{figure}

We also investigated the prediction latency of SNNs trained using different methods. This corresponds to the amount of simulation time needed by the models to reach full confidence in their predictions and determines the simulation duration required to achieve high accuracy.
\par
Figure \ref{fig:prediction_confidence} shows the averaged prediction confidence over time for SNNs trained using each method. We found that, when initialized with only positive weights, Fast \& Deep achieved high confidence on predictions after more than 150ms. However, when initialized with both negative and positive weights, both Fast \& Deep and our proposed method achieved confidence earlier (in 20ms and 50ms, respectively). This suggests that initialization has a significant impact on prediction latency. However, SNNs trained with our proposed method are slightly slower than TTFS networks trained with Fast \& Deep for the same weight distribution due to the increased spike count per neuron.
\par
Despite this difference, the short latency of these networks allows for shorter simulation durations, which can further improve sparsity without affecting performance. In Figure \ref{fig:spike_count_vs_accuracy}, we show the relationship between population spike count and accuracy for each SNN as simulation time increases from 0 to 200 milliseconds. This demonstrates that, by reducing the simulation time, SNNs can become more sparse while maintaining high performance. Therefore, the sparsity-accuracy tradeoff can be further improved after training by adjusting the simulation duration. This also demonstrates that our method can align with the level of the sparsity of Fast \& Deep while still performing better.

\subsection{Robustness to Noise and Weights Quantization}

\begin{figure}[!tbp]
	\centering
	\begin{subfigure}[t]{0.45\textwidth}
		\centering
		\begin{tikzpicture}[trim axis left, trim axis right]
			\begin{axis}[
				xlabel={Jitter standard deviation $\sigma$ (ms)},
				ylabel style={align=left}, 
				ylabel={Normalized acc. (\%)},
				legend style={
					at={(-0.3, 1.3)}, 		
					anchor=west,
					cells={align=center},
					/tikz/every even column/.append style={column sep=0.5cm},
					legend columns=-1
				},
				width=1.0\linewidth,
				height=0.8\linewidth,
				ymajorgrids=true,
				grid style=dashed]
				
				\addplot[greenmat, thick, mark=triangle, mark options={solid}, densely dotted] coordinates {
					(0, 96.73 * 100 / 96.73)(1, 96.49 * 100 / 96.73) (2, 96.31 * 100 / 96.73) (3, 96.2 * 100 / 96.73)
				};
				\addplot[bluemat, thick, mark=square, mark options={solid}, dashed] coordinates {
					(0, 97.73 * 100 / 97.75)(1, 97.75 * 100 / 97.75)(2, 97.45 * 100 / 97.75)(3, 97.36 * 100 / 97.75)
				};
				\addplot[redmat, thick, mark=*] coordinates {
					(0, 98.89 * 100 / 98.92)(1, 98.92 * 100 / 98.92)(2, 98.84 * 100 / 98.92)(3, 98.87 * 100 / 98.92)
				};
				\legend{{Fast \& Deep $U(-1, 1)$}, {Fast \& Deep $U(0, 1)$}, {Our method $U(-1, 1)$}}
			\end{axis}
		\end{tikzpicture}
		\caption{}
		\label{fig:spike_jitter}
	\end{subfigure}%
	\begin{subfigure}[t]{0.45\textwidth}
		\centering
		\begin{tikzpicture}[trim axis left, trim axis right]
			\begin{axis}[
				xlabel={$w_{\text{clip}}$},
				ylabel style={align=left},
				ylabel={Normalized acc. (\%)},
				width=1.0\linewidth,
				height=0.8\linewidth,
				ymajorgrids=true,
				grid style=dashed]
				
				\addplot[greenmat, thick, mark=triangle, mark options={solid}, densely dotted] coordinates {
					(0.5, 96.71 * 100 / 96.86) (1.0, 96.86 * 100 / 96.86) (1.5, 96.86 * 100 / 96.86)(2, 96.32 * 100 / 96.86)
				};
				\addplot[bluemat, thick, mark=square, mark options={solid}, dashed] coordinates {
					(0.5, 94.48 * 100 / 97.83)(1.0, 97.16 * 100 / 97.83)(1.5, 97.83 * 100 / 97.83)(2.0, 97.83 * 100 / 97.83)
				};
				\addplot[redmat, thick, mark=*] coordinates {
					(0.5, 97.83 * 100 / 98.7) (1.0, 98.5 * 100 / 98.7) (1.5, 98.65 * 100 / 98.7) (2.0, 98.7 * 100 / 98.7)
				};
			\end{axis}
		\end{tikzpicture}
		\caption{}
		\label{fig:weight_clipping}
	\end{subfigure}
	\begin{subfigure}[t]{0.45\textwidth}
		\centering
		\begin{tikzpicture}[trim axis left, trim axis right]
			\begin{axis}[
				xlabel={$n$ (bit)},
				ylabel style={align=left}, 
				ylabel={Normalized acc. (\%)},
				legend style={at={(0.03,0.5)},anchor=west},
				xtick={2, 3, 4, 5, 6},
				xticklabels={2, 3, 4, 5, float},
				width=1.0\linewidth,
				height=0.8\linewidth,
				ymajorgrids=true,
				grid style=dashed]
				
				\addplot[greenmat, thick, mark=triangle, mark options={solid}, densely dotted] coordinates {
					(2, 95.79 * 100 / 96.86)(3, 95.83 * 100 / 96.86) (4, 96.01 * 100 / 96.86) (5, 96.06 * 100 / 96.86) (6, 96.86 * 100 / 96.86)
				};
				\addplot[bluemat, thick, mark=square, mark options={solid}, dashed] coordinates {
					(2, 96.49 * 100 / 97.16)(3, 96.91 * 100 / 97.16) (4, 96.95 * 100 / 97.16) (5, 97.1 * 100 / 97.16)(6, 97.16 * 100 / 97.16)
				};
				\addplot[redmat, thick, mark=*] coordinates {
					(2, 98.34 * 100 / 98.5)(3, 98.38 * 100 / 98.5) (4, 98.4 * 100 / 98.5) (5, 98.5 * 100 / 98.5) (6, 98.5 * 100 / 98.5)
				};
			\end{axis}
		\end{tikzpicture}
		\caption{}
		\label{fig:weight_precision}
	\end{subfigure}
	\caption{Figure \ref{fig:spike_jitter} shows the effect of spike jitter on the performance of each method. This was achieved by introducing artificial noise to the spike timings, following a normal distribution $\mathcal{N}(0, \sigma)$. Figure \ref{fig:weight_clipping} displays the impact of weight clipping, which involved restricting weights to the range $\left[-w_{\text{clip}}, w_{\text{clip}}\right]$ during training. Lastly, Figure \ref{fig:weight_precision} demonstrates the effect of weight precision, which was obtained by discretizing weights into $2^{n+1} - 1$ bins ($n$ bits plus one bit for the sign of the synapse) within the range $\left[-1,1\right]$. Overall, our method was found to be more resilient to noise and reduced weight precision than Fast \& Deep.}
	\label{fig:robustness_noise}
	\vspace{0.2in}
\end{figure}
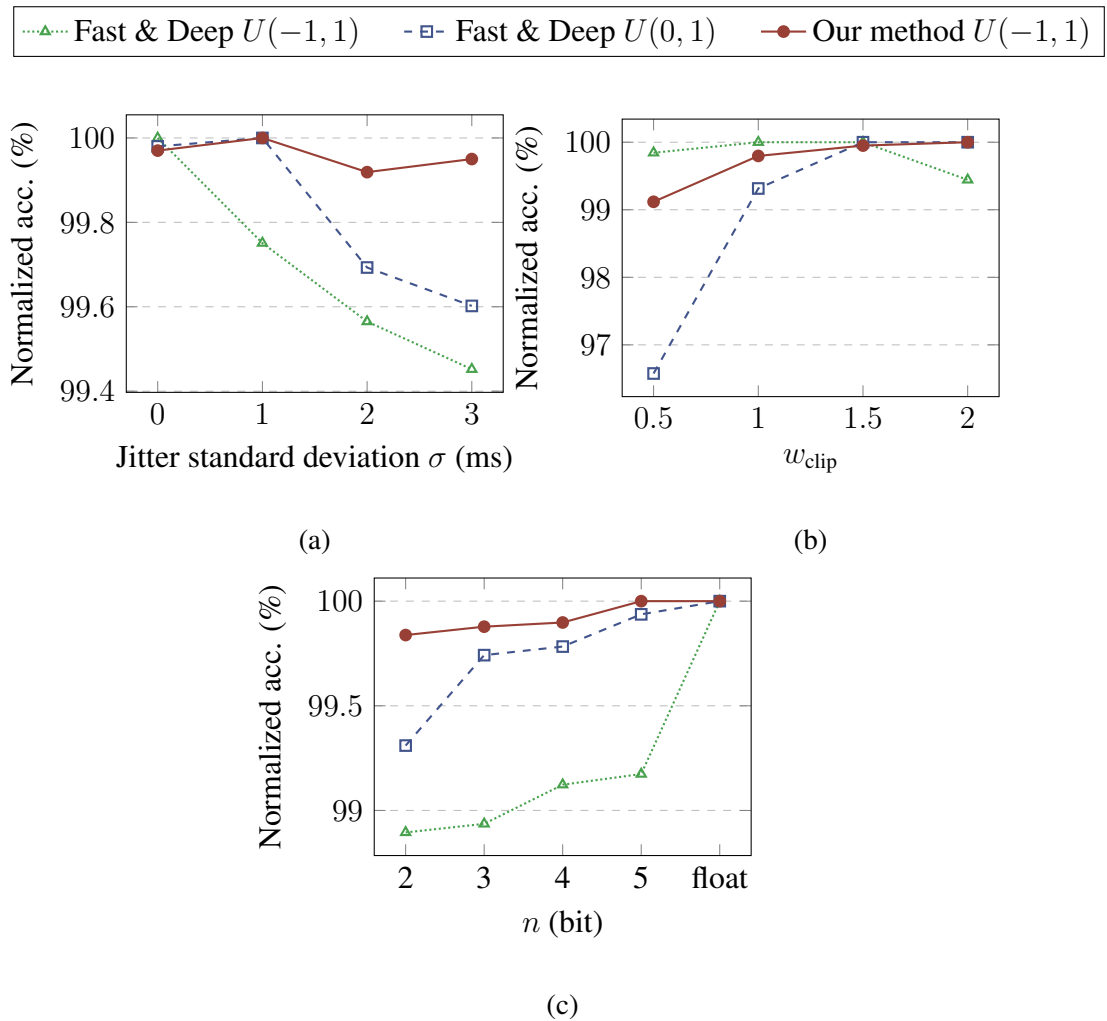

Analog neuromorphic hardware are inherently noisy and are often limited to specific ranges and resolutions of weights. Having a model that is robust to noise and weight quantization is therefore important to achieve high performance on such systems. To assess the robustness of each method, we measured the impact of spike jitter, weight clipping, and weight precision on their accuracy. In these experiments, performance was normalized with the maximum accuracy of each method to better compare variations. 
\par
Figure \ref{fig:spike_jitter} shows the impact of spike jitter on the performance of each method. These results were produced by artificially adding noise to spike timings with a normal distribution $\mathcal{N}(0, \sigma)$ during training. When initialized with both negative and positive weights, Fast \& Deep appears to be less robust than using only positive weights. With negative weights, only a fraction of neurons transmits information which leads to an increased sparsity, as illustrated in Figure \ref{fig:spike_count_active}. Therefore, introducing noise to spike timings significantly impacts performance. In contrast, positive weights ensure consistent network activity and redundancy in transmitted information. SNNs initialized with positive weights are thus less affected by spike jitter. However, SNNs trained with Fast \& Deep remain susceptible to noise, even with positive initial weights. Perturbations in spike timing still have a critical impact on temporal coding. In contrast, our proposed method demonstrates greater robustness to spike jitter than Fast \& Deep, with minimal variation observed. This is a result of the redundancy created by the multiple spikes fired by neurons.
\par
In Figure \ref{fig:weight_clipping}, we demonstrate the effect of weight range on performance by clipping weights between the range $\left[-w_{\text{clip}}, w_{\text{clip}}\right]$ during training. The performance of Fast \& Deep initialized with positive weights degrades when $w_{\text{clip}}$ is lower than 1.5. However, both our method and Fast \& Deep exhibit robustness to reduced weight ranges when initialized with both negative and positive weights. This suggests that weight distribution may play a role in the network's resilience to limited weight ranges. 
\par
Finally, Figure \ref{fig:weight_precision} shows the performance of each method with reduced weight resolutions from 5 to 2 bits (results with float precision are also given as a reference). It highlights that Fast \& Deep is less robust to reduced weight precision than our method, particularly with negative weights. In contrast, our approach is only slightly impacted by the decreased precision, even when reduced to as low as 2 bits.

\section{Discussion}

In this work, we explored the tradeoffs between performance and various aspects of TTFS SNNs such as sparsity, classification latency and robustness to noise and weight quantization. We also generalized the Fast \& Deep algorithm by incorporating a reset of the membrane potential which enables multiple spikes per neuron and compared the improvements of the proposed method with the origin algorithm on those tradeoffs.
\par
We found that initializing Fast \& Deep with positive weights leads to better generalization capabilities compared to initializing with both negative and positive weights. This observation is consistent across the benchmarked datasets, as shown in Table \ref{table:performance_comparison}. However, relaxing the spike constraint improves the overall performance and convergence rate of SNNs, at least on the benchmark problems we considered. This result was expected before our experiments since BP methods that use multiple spikes per neuron generally perform better than methods that impose firing constraints \citep{macro_micro_backpropagation, spike_train_level_backpropagation, slayer, training_deep_snn_using_bp, STDBP}.
\par
Our experiments also demonstrate that the weight distribution significantly influences the sparsity of Fast \& Deep. We observed that SNNs with positive weight initialization tend to be less sparse than those initialized with weights between -1 and 1. However, the former consistently outperforms the latter in terms of performance. This highlights the accuracy-sparsity tradeoff often observed when training SNNs \citep{performance_sparsity_tradeoff, rethinking_gradiet_descent_for_snn}. The quasi-dense activity provided by positive weights explains the difference in sparsity, as shown in Figure \ref{fig:activity_fast_and_deep_01}. In contrast, initializing Fast \& Deep with both negative and positive weights leads to fewer active neurons due to the inhibition provided by negative weights. Additionally, neurons trained with Fast \& Deep fire indiscriminately to stimuli, suggesting a pure temporal representation of information, whereas neurons trained with our proposed method selectively respond to their inputs and exhibit a different distribution of activity, as shown in Figure \ref{fig:activity_bats}. Our unconstrained SNNs allow for a different distribution of the spike activity, whereby key neurons can fire more often than others, while irrelevant neurons may not spike at all. This enables our method to achieve a level of sparsity comparable to Fast \& Deep on image classification, as illustrated in Figure \ref{fig:spike_count_active}.
\par
To achieve a high degree of sparsity without firing rate regularization, thresholds can be tuned to indirectly influence spiking activity through learning. Decreasing thresholds increases the firing rate of downstream layers, resulting in more negative errors at outputs and consequently negative weight changes in hidden layers, as shown in Figure \ref{fig:cost_vs_accuracy}. This mechanism offers a natural way to control sparsity in unconstrained SNNs without requiring any regularization technique. By leveraging this principle, we were able to train shallow SNNs with a sparsity level similar to Fast \& Deep while achieving higher performance. This implies that allowing multiple spikes per neuron has the potential to enhance the accuracy-sparsity tradeoff and prompts further investigation into the effectiveness of TTFS in achieving efficient computation. However, finding thresholds that lead to high sparsity is more difficult when networks become deeper due to the fluctuations in the firing rates of each layer. The factors that influence sparsity in unregularized SNNs are currently not fully understood and present an opportunity for future research to investigate how to naturally achieve sparsity in deep architectures.
\par
Our experiments on the SHD dataset also demonstrated the significance of the accuracy-sparsity tradeoff when processing temporal data. We found that the ability of neurons to fire multiple spikes is critical in capturing all the information about the inputs over time. However, since TTFS neurons encode important information in early spikes, they tend to fire too early to capture all the information, which makes them less effective than unconstrained SNNs on temporal data. Although TTFS SNNs are more energy-efficient, the relaxation of spike constraints in unconstrained SNNs allows neurons to fire throughout the simulation, thereby capturing all the relevant information. Consequently, they perform significantly better than TTFS SNNs when processing temporal data.
\par
In addition to performance and sparsity, we measured the prediction latency of each method, which is the waiting time required before the system can make reliable predictions. We found that the speed of classification was primarily driven by the weight distribution. Figure \ref{fig:prediction_confidence} shows that both Fast \& Deep and our method achieve similar latencies when initialized between -1 and 1, with a slight advantage for Fast \& Deep. However, when initialized with only positive values, Fast \& Deep requires more simulation time to achieve full confidence in predictions. Low latency is advantageous not only in terms of inference speed but also in improving energy efficiency. If predictions occur early enough, the duration of simulations can be significantly reduced, limiting the number of spikes fired by neurons. In Figure \ref{fig:spike_count_vs_accuracy}, we demonstrated that reducing the simulation time can lead to a reduction in computational cost for both Fast \& Deep and our method while maintaining the same performance. This shows that prediction latency, energy consumption, and performance are closely related and that unconstrained SNNs can offer better tradeoffs between these aspects than TTFS SNNs.
\par
The last characteristic that we investigated is the robustness to noise and weight quantization that are inherent to analog neuromorphic hardware. The timing of spikes is a critical factor for the performance of TTFS SNNs as it carries most of the information. Therefore, perturbations in these timings and weight constraints can significantly affect the reliability of the feature extraction. In contrast, our proposed method benefits from an increased number of spikes per neuron, providing redundancy that enhances resilience to noise and weight constraints. For instance, Figure \ref{fig:spike_jitter} demonstrates that our method is less impacted by perturbations in spike timings than Fast \& Deep. This suggests that our proposed method has the potential to provide more stable learning on analog neuromorphic hardware than Fast \& Deep.
\par
Finally, our work specifically concentrated on backpropagation in SNNs with fixed thresholds and time constants. However, several studies have demonstrated that incorporating adaptive thresholds and trainable time constants can enhance the convergence, sparsity, and performance of SNNs \citep{sparse_adaptive_snn, adaptive_threshold_deep_csnn, learnable_membrane_time_constant, adaptive_time_constant_rsnn}. Therefore, future studies could explore the integration of adaptive thresholds and trainable time constants into our proposed method to further enhance the sparsity-accuracy tradeoffs in SNNs.

\section{Conclusion}

Our work demonstrates that relaxing the spike constraint of TTFS SNNs results in improved tradeoffs among performance, sparsity, latency, and noise robustness. Our findings also highlight the crucial role of thresholds in regulating the sparsity of unconstrained SNNs during learning, which could serve as a natural alternative to firing rate regularization. Although error backpropagation algorithms for SNNs are incompatible with neuromorphic hardware, their development provides valuable insights into how spiking neurons affect objective functions and could support the development of hardware-compatible algorithms. Therefore, our work contributes to a better understanding of how to compute exact gradients in SNNs and highlights the advantages of using multiple spikes per neuron over TTFS.

\section{Code Availability}

The code produced in this work will be made available at: https://github.com/Florian-BACHO/bats

\section*{Appendix}

\subsection*{Experimental Settings}

\subsubsection*{Simulations}

We implemented the method introduced in Section \ref{section:method} in a custom Python simulator for GPUs using CuPy \citep{cupy}. In our implementation, both simulations and error backpropagation are event-based.

\subsubsection*{Input Encoding}

We used a TTFS encoding scheme to benefit from a low number of input spikes and fast processing. For image classification tasks, we encoded the pixels into spike timing as follows:
\par 
Given the time window $T_{\text{enc}} = 100$ milliseconds and the maximum pixel value $X_{\text{max}} = 255$, the input spike timing $t^{(1,i j_{\text{max}} + j)}_1$ associated with the pixel value $x_{i,j}$ in row $i$ and column $j$ is computed as:
\begin{equation}
	t^{(1,i j_{\text{max}} + j)}_1 = \frac{T_{\text{enc}}}{X_{\text{max}}} \left(X_{\text{max}} - x_{i,j}\right)
\end{equation}
where $j_{\text{max}}$ is the width of the image in pixels.
Neurons with a pixel value $x_{i,j} = 0$ do not produce any spikes to further limits the number of events to process. For convolutional SNNs, the same temporal encoding was used but the shape of the input was set as a three-dimensional tensor corresponding to the image with a single channel.

\subsubsection{Implementation of Fast \& Deep}

To reproduce Fast \& Deep with TTFS models, we constrained the number of firing allowed per neuron to one in our implementation and used a TTFS softmax cross-entropy loss function, as described in \citep{deep_fast}. 

\subsubsection{Architectures and Parameters}

For the MNIST EMNIST and SHD datasets, we trained both TTFS and unconstrained fully-connected SNNs with a batch size of 50 and a maximum number of spikes per neuron of 30 for the unconstrained SNNs. Output spike targets were set to 15 for the target label and 3 for the others. We used a learning rate of $\lambda=0.003$ for image classification and $\lambda=0.001$ for the SHD dataset. No data augmentation was used with full connected networks.
\par
For Fashion-MNIST, we implemented a three-layer fully-connected network composed of two hidden layers of 400 neurons each and a 10 neuron output layer. We allowed a maximum number of spikes per neuron of 5 for the hidden layers and 20 for the output layer. We also set the target spike counts to 15 for the true class and 3 for the others. We used a batch size of 5 with a learning rate of $\lambda=0.0005$, a learning rate decay factor of $0.5$ every 10 epochs and a minimum rate of 0.0001.
\par
The weight kernels of convolution neurons were shared within each layer, as in rate-based CNNs. Convolution allows the detection of spatially-correlated features and therefore, makes networks invariant to translations. In contrast to fully connected SNNs, the translations invariance of CSNNs allows the networks to detect objects at different locations in space. We used a 6 layer CSNN composed of two spiking convolutional layers of 15 5x5 and 40 5x5 filters respectively, each followed by 2x2 spike aggregation pooling layers (i.e. the spike trains of input neurons are aggregated into a single spike train). The spikes of the last pooling layer are finally sent to two successive fully-connected layers of sizes 300 and 10 respectively. Each layer allows an increasing number of spikes per neuron, starting from a single spike for the first convolutional layer, 3 for the second layer, 10 for the fully-connected layer and 30 spikes per neuron for the output layer. We also set the output spike targets to 30 for the true label and 3 for the others. The CSNN was also trained with data augmentation. In this case, we used Elastic Distortions \cite{elastic_distortion} to transform the MNIST training images. We finally trained the networks for 100 epochs with a batch size of 20, a learning rate of $\lambda=0.003$, a decay factor of 0.5 every 10 epochs and a minimum rate of 0.0001.
\par
In all our experiments, we used the Adam optimizer with the values of $\beta_1$, $\beta_2$ and $\epsilon$ set as in the original paper \citep{adam}. Initial weights were randomly drawn from a uniform distribution $U[a,b]$. Networks trained on image classification used the same base time constant of $\tau_s=0.130$. For the SHD dataset, we used a time constant of $\tau_s=0.100$. All thresholds were manually tuned to find the best-performing networks for each method and dataset. Thresholds were then kept fixed during training. Finally, we did not use any regularization or synaptic scaling techniques in any of our experiments to provide a fair comparison between TTFS and unconstrained SNNs.

\subsection*{Fully-connected SNNs on MNIST}

\begin{table}[H]
	\centering
	\caption{Performances of several methods on the MNIST dataset. Results for Fast \& Deep and our method are highlighted in bold.}
	\label{table:mnist_performances}
	\begin{center}
		\begin{tabular}{|c|lll|}
			\hline
			& Method & Arch. & Test accuracy \\ \hline
			\parbox[t]{4mm}{\multirow{9}{*}{\rotatebox[origin=c]{90}{TTFS}}} & Fast \& Deep \citep{deep_fast} & 350  & 97.1 $\pm$ 0.1\% \\
			& Wunderlich \& Pehle \citep{event_based_exact_gradient_snn} & 350 & 97.6 $\pm$ 0.1\% \\
			& Alpha Synapses \citep{gradient_descent_alpha_function} & 340 & 97.96\% \\
			& S4NN \citep{S4NN} & 400 & 97.4 $\pm$ 0.2\% \\
			& BS4NN \citep{BS4NN} & 600 & 97.0\% \\
			& Mostafa \citep{supervised_learning_based_on_temporal_coding} & 800 & 97.2\% \\
			& STDBP \citep{STDBP} & 800 & 98.5\% \\
			& \begin{tabular}[c]{@{}l@{}}
				\textbf{Fast \& Deep \citep{deep_fast}}\\
				\textbf{(our implementation)}  \end{tabular} & \textbf{800} & \textbf{97.83} $\boldsymbol{\pm}$ \textbf{0.08\%} \\ \hline
			\parbox[t]{4mm}{\multirow{4}{*}{\rotatebox[origin=c]{90}{Unconstrained}}} & eRBP \citep{event_driven_random_bp} & 2x500 & 97.98\% \\
			& Lee et al. \citep{training_deep_snn_using_bp} & 800 & 98.71\% \\
			& HM2-BP \citep{macro_micro_backpropagation} & 800 & 98.84 $\pm$ 0.02\% \\
			& \textbf{This work} & \textbf{800} & \textbf{98.88} $\boldsymbol{\pm}$ \textbf{0.02}\% \\ \hline
		\end{tabular}
	\end{center}
\end{table}

\subsection*{Fully-connected SNNs on EMNIST}

\begin{table}[H]
	\centering
	\caption{Performances of several methods on the EMNIST dataset. Results for Fast \& Deep and our method are highlighted in bold.}
	\label{table:emnist_performances}
	\begin{center}
		\begin{tabular}{|c|lll|}
			\hline
			& Method & Arch. & Test accuracy \\ \hline
			\parbox[t]{4mm}{\multirow{1}{*}{\rotatebox[origin=c]{90}{TTFS}}} & \begin{tabular}[c]{@{}l@{}}
				\textbf{Fast \& Deep \citep{deep_fast}} \\
				\textbf{(our implementation)}  \end{tabular} & \textbf{800} & \textbf{83.34} $\boldsymbol{\pm}$ \textbf{0.27\%} \\ \hline
			\parbox[t]{4mm}{\multirow{4}{*}{\rotatebox[origin=c]{90}{Unconstrained}}} & eRBP \citep{event_driven_random_bp} & 2x200 & 78.17\% \\
			& HM2-BP \citep{macro_micro_backpropagation} & 2x200 & 84.31 $\pm$ 0.10\% \\
			& HM2-BP \citep{macro_micro_backpropagation} & 800 & 85.41 $\pm$ 0.09\% \\
			& \textbf{This work} & \textbf{800} & \textbf{85.75 $\boldsymbol{\pm}$ 0.06\%} \\
			\hline
		\end{tabular}
	\end{center}
\end{table}

\subsection*{Fully-connected SNNs on Fashion MNIST}

\begin{table}[H]
	\centering
	\caption{Performances of several methods on the Fashion-MNIST dataset. Results for Fast \& Deep and our method are highlighted in bold. * means that the trained model has recurrent connections.}
	\label{table:fashion_mnist_performances}
	\begin{center}
		\begin{tabular}{|c|lll|}
			\hline
			& Method & Architecture & Test accuracy \\ \hline
			\parbox[t]{4mm}{\multirow{5}{*}{\rotatebox[origin=c]{90}{TTFS}}} & S4NN \citep{S4NN} & 1000 & 88.0\% \\
			& BS4NN \citep{BS4NN} & 1000 & 87.3\% \\
			& STDBP \citep{STDBP} & 1000 & 88.1\% \\ 
			& \begin{tabular}[c]{@{}l@{}}
				\textbf{Fast \& Deep \citep{deep_fast}}\\
				\textbf{(our implementation)}  \end{tabular} & \textbf{2x400} & \textbf{88.28} $\boldsymbol{\pm}$ \textbf{0.41\%} \\ \hline
			\parbox[t]{4mm}{\multirow{4}{*}{\rotatebox[origin=c]{90}{Unconstrained}}} & HM2-BP \citep{macro_micro_backpropagation} & 2x400 & 88.99\% \\
			& TSSL-BP \citep{temporal_spike_sequence_learning} & 2x400 & 89.75 $\pm$ 0.03\% \\
			& ST-RSBP* \citep{spike_train_level_backpropagation} & 2x400 & 90.00 $\pm$ 0.14\% \\
			& \textbf{This work} & \textbf{2x400} & \textbf{90.19 $\boldsymbol{\pm}$ 0.12\%} \\
			\hline
		\end{tabular}
	\end{center}
\end{table}

\subsection*{Fully-connected SNNs on SHD}

\begin{table}[H]
	\centering
	\caption{Performances of several methods on the Spiking Heidelberg Digits (SHD) dataset. Results for Fast \& Deep and our method are highlighted in bold. * means that the trained model has recurrent connections.}
	\label{table:shd_performances}
	\begin{center}
		\begin{tabular}{|c|lll|}
			\hline
			& Method & Architecture & Test accuracy \\ \hline
			\parbox[t]{4mm}{\multirow{1}{*}{\rotatebox[origin=c]{90}{TTFS}}}
			& \begin{tabular}[c]{@{}l@{}}
				\textbf{Fast \& Deep \citep{deep_fast}}\\
				\textbf{(our implementation)}  \end{tabular} & \textbf{128} & \textbf{47.37} $\boldsymbol{\pm}$ \textbf{1.65\%} \\ \hline
			\parbox[t]{4mm}{\multirow{3}{*}{\rotatebox[origin=c]{90}{Unconstrained}}} & Cramer et al. \citep{heidelberg_dataset} & 128 & 48.10 $\pm$ 1.6\% \\
			& Cramer et al.* \citep{heidelberg_dataset} & 128 & 71.4 $\pm$ 1.9\% \\
			& \textbf{This work} & \textbf{128} & \textbf{66.79 $\boldsymbol{\pm}$ 0.66\%} \\
			\hline
		\end{tabular}
	\end{center}
\end{table}

\subsection*{Convolutional SNNs on MNIST}

\begin{table}[H]
	\centering
	\caption{Network architectures used in Table \ref{table:cnn_mnist_performances}. 15C5 represents a convolution layer with 15 5x5 filters and P2 represents a 2x2 pooling layer.}
	\label{table:cnn_architectures}
	\begin{center}
		\begin{tabular}{|l|l|}
			\hline
			Network Name & Architecture \\ \hline
			Net1 & 32C5-P2-16C5-P2-10 \\
			Net2 & 12C5-P2-64C5-P2-10 \\
			Net3 & 15C5-P2-40C5-P2-300-10 \\
			Net4 & 20C5-P2-50C5-P2-200-10 \\
			Net5 & 32C5-P2-32C5-P2-128-10 \\
			Net6 & 16C5-P2-32C5-P2-800-128-10 \\
			\hline
		\end{tabular}
	\end{center}
\end{table}

\begin{table}[H]
	\centering
	\caption{Performances of several methods on the MNIST dataset with Spiking Convolutional Neural Networks. The network topologies are given in Table \ref{table:cnn_architectures}. * means that the network has been trained using data augmentation.}
	\label{table:cnn_mnist_performances}
	\begin{center}
		\begin{tabular}{|c|lll|}
			\hline
			& Method & Arch. & Test accuracy \\ \hline
			\parbox[t]{4mm}{\multirow{4}{*}{\rotatebox[origin=c]{90}{TTFS}}} & Zhou et al* \citep{temporal_coded_deep_snn} & Net1 & 99.33\% \\
			& STDBP* \citep{STDBP} & Net6 & 99.4\% \\ 
			& \textbf{Fast \& Deep (our implementation)} & \textbf{Net3} & \textbf{99.22} $\boldsymbol{\pm}$ \textbf{0.05\%} \\
			& \textbf{Fast \& Deep* (our implementation)} & \textbf{Net3} & \textbf{99.46} $\boldsymbol{\pm}$ \textbf{0.01\%} \\
			\hline
			\parbox[t]{4mm}{\multirow{7}{*}{\rotatebox[origin=c]{90}{Unconstrainned}}} & Lee et al.* \citep{training_deep_snn_using_bp} & Net4 & 99.31\% \\
			& HM2-BP* \citep{macro_micro_backpropagation} & Net3 & 99.42\% $\pm$ 0.11\% \\
			& TSSL-BP \citep{temporal_spike_sequence_learning} & Net3 & 99.50 $\pm$ 0.02\% \\
			& ST-RSBP* \citep{spike_train_level_backpropagation} & Net2 & 99.50 $\pm$ 0.03\% \\
			& ST-RSBP* \citep{spike_train_level_backpropagation} & Net3 & 99.57 $\pm$ 0.04\% \\
			& \textbf{This work} & \textbf{Net3} & \textbf{99.38} $\boldsymbol{\pm}$ \textbf{0.04\%} \\
			& \textbf{This work*} & \textbf{Net3} & \textbf{99.60 $\boldsymbol{\pm}$ 0.03\%} \\
			\hline
		\end{tabular}
	\end{center}
\end{table}

\bibliographystyle{APA}
\bibliography{bibliography}

\end{document}